\RequirePackage{fix-cm}

\documentclass[smallextended,numbook]{svjour3}

\usepackage{amsmath,amssymb,amsfonts}
\usepackage{mathtools}
\usepackage{graphicx}
\usepackage{tikz}
\usepackage{textcomp}
\usepackage{booktabs}
\usepackage{xcolor}
\def\BibTeX{{\rm B\kern-.05em{\sc i\kern-.025em b}\kern-.08em
    T\kern-.1667em\lower.7ex\hbox{E}\kern-.125emX}}
\usepackage[hyphens]{url} 
\usepackage{tikz}
\usepackage{microtype}
\usepackage{natbib}

\usepackage[normalem]{ulem} 
\newcommand{\added}[2]{#2}
\newcommand{\replaced}[3]{#2} 

\usepackage[utf8]{inputenc}

\smartqed

\usepackage{cleveref}
\let\cref\Cref

\begin{document}
\title{Topic Space Trajectories}
\subtitle{A Case Study on Machine Learning Literature}

\author{Bastian Schaefermeier \and
  Gerd Stumme \and
  Tom Hanika
}

\institute{B. Schaefermeier \at
              	L3S Research Center, Appelstra\ss e 9a, Hannover, Germany \\
		Tel.: +49 561 804-6254\\
		Fax.: +49 561 804-6259\\
              \email{bsc@cs.uni-kassel.de}           
           \and
           Gerd Stumme \at
              Knowledge and Data Engineering Group, University of Kassel, Wilhelmshoeher Allee 73, Kassel, Germany\\
              \email{stumme@cs.uni-kassel.de}
              \and
           Tom Hanika \at
              Knowledge and Data Engineering Group, University of Kassel, Wilhelmshoeher Allee 73, Kassel, Germany\\
              \email{tom.hanika@cs.uni-kassel.de}
}

\date{Received: date / Accepted: date}

\maketitle
\begin{abstract} 
  The annual number of publications at scientific venues, for example,
  conferences and journals, is growing quickly. Hence, even for researchers it
  becomes harder and harder to keep track of research topics
  and their progress. In this task, researchers can be supported by automated
  publication analysis. Yet, many such methods result in
  uninterpretable, purely numerical representations.

  As an attempt to support human analysts, we present \emph{topic
    space trajectories}, a structure that allows for the
  comprehensible tracking of research topics. We demonstrate how these
  trajectories can be interpreted based on eight different analysis
  approaches.  To obtain comprehensible results, we employ
  non-negative matrix factorization as well as suitable visualization
  techniques. We show the applicability of our approach on a
  publication corpus spanning 50 years of machine learning research
  from 32 publication venues. \added{Review 1.8}{In addition to a thorough introduction of our method, our focus is on an extensive analysis of the results we achieved.} Our novel analysis method may be
  employed for paper classification, for the prediction of future
  research topics, and for the recommendation of fitting conferences
  and journals for submitting unpublished work. \added{Review 2.1}{An advantage in these applications over previous methods
  lies in the good interpretability of the results obtained through our methods.}
\end{abstract}

\setcounter{tocdepth}{3} 


\section{Introduction}
The number of publications published in scientific venues, such as
journals and conferences, is vastly increasing. For instance, there
were at least 5000 papers at major machine learning conferences and
journals in 2018, more than twice as many as in
2008~\citep{semanticscholar}. Even though many venues emphasize a
particular research field they do exhibit a plurality of topics. This
is a natural consequence observed when research fields grow, which
leads to new specializations and the emergence of new research
topics. The related advance in knowledge is, however, overshadowed by
the increasing inability to maintain a comprehensive overview of a
research field. Moreover, even for experts understanding and tracking
the topic dynamics of contemporary research fields is an infeasible
endeavor. Hence, the need for an automated approach to cope with the
aforesaid vast amount of publication data is pressing.

Answering to this we present a novel conceptualization for topic based analysis
of scientific publication venues. Our approach is based on \emph{topic models},
a common class of methods to analyze text corpora. By computing topic vectors
for research papers we are able to position the associated publication venues in
a low-dimensional topic space. This enables the comparison of conferences as well as journals and
following their temporal (topical) dynamics. A governing constraint in our work is
to compute human interpretable results. Thus, we employ non-negative matrix
factorization (NMF), a model whose topics are comparatively well comprehensible.
In detail, we aggregate document representations calculated by NMF over
conferences/journals and years. This allows to capture the temporal
topical development of scientific venues. Moreover, we semi-automatically select an
appropriate number of topics based on a coherence measure. Additionally we
take advantage of proper visualization techniques.

Altogether this results in
the scientometric analysis of what we call \emph{topic space trajectories}
(TST). Despite the availability of almost all the building blocks, no one, to the
best of our knowledge, has presented an akin notion for publication data
analysis.
We demonstrate the applicability of our approach on a publication data
set of the top tier machine learning conferences covering the years
from 1969 to 2018. In particular we track 32 publication venues in
topic space and depict their topical drift. What is more, we introduce
topic densities (i.e., distributions) for publication venues.
Besides scientometric analysis we envision multiple applications of our
work. First, we think that TST may be employed in conference or reviewer recommendations
for new research papers in classification based approaches.
Second, one may extrapolate the topical drift of conferences into the future.
Third, our method may be transferred to the analysis of other text
domains, such as news articles or patent applications.

Our work is organized as follows: First, we present our method for
obtaining document and venue representations as well as their
trajectories in topic space (\cref{sec:tst}). In the
subsequent~\cref{sec:experiments}, we introduce our case study on a
machine learning publication corpus. We demonstrate for
the resulting topic space a collection of interpretation approaches
in~\cref{sec:interpret}. Before concluding our work
in~\cref{sec:conclusion} we give an overview of related work
in~\cref{sec:related}.

\section{Topic Space Trajectories}\label{sec:tst}
In this section we describe our overall method, which is depicted in
Figure~\ref{theplot}. The essential part is the publication corpus,
from which we extract a
subset of papers relevant for the analysis, for example, the subset of
machine learning papers of certain venues. From this corpus we train a
topic model using non-negative matrix factorization
(NMF). Additionally to topics, we obtain embeddings (i.e.,
representations) of papers and venues, which will allow the definition
of venue trajectories in topic space.

\usetikzlibrary{shapes,arrows} 
\tikzstyle{proc} = [rectangle, draw, minimum height=2.5em, node distance=3cm]
\tikzstyle{res} = [minimum height=2em, node distance=7cm, text width=5.5em, text centered]
\tikzstyle{line} = [draw, -latex',>=stealth, shorten <=2pt,shorten >=2pt]

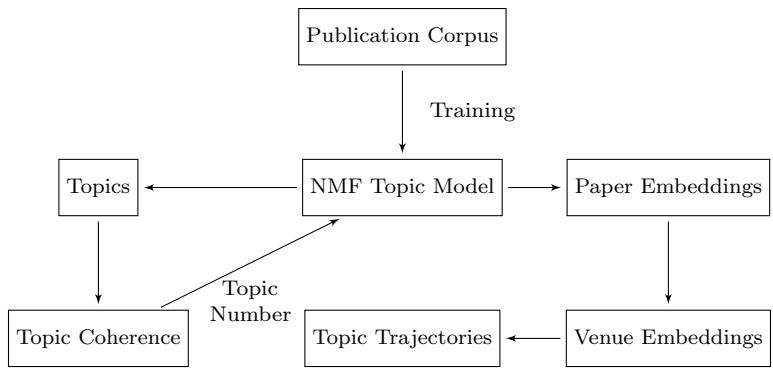
\begin{figure}
  \begin{center}
    \begin{tikzpicture}[node distance=2.8cm,auto]
\node [proc, node distance=5cm] (mlc) {Publication Corpus};
\node [proc, below of=mlc, node distance=2cm] (nmf) {NMF Topic Model};
\node [proc, left of=nmf, node distance=4cm] (topics) {Topics};
\node [proc, right of=nmf, node distance=3.5cm] (embed) {Paper Embeddings};
\node [proc, below of=embed, node distance=2cm] (venues) {Venue Embeddings};
\node [proc, below of=nmf, node distance=2cm] (trajectories) {Topic Trajectories};
\node [proc, below of=topics, node distance=2cm] (coherence) {Topic Coherence};

\path [line] (mlc) -- node[res]{Training}(nmf);
\path [line] (nmf) -- node[res]{}(topics);
\path [line] (nmf) -- node[res]{}(embed);
\path [line] (venues) -- node[res]{}(trajectories);
\path [line] (embed) -- node[res]{}(venues);
\path [line] (topics) -- node[res]{}(coherence);

\path [line] (coherence) -- node[res, below of=nmf, node distance=0.5cm]{Topic Number}(nmf);

\end{tikzpicture}
  \end{center}
\caption{Depiction of our method. Each box contains some data or (intermediate) result. Arrows indicate where data or results are an input for the subsequent box.}
\label{theplot}
\end{figure}

\subsection{Publication Corpus}
We consider a corpus of publications $\mathcal{D}$ as a set of papers,
each of which appeared at a certain venue (conference or journal) in a
certain year.  At this point we omit a thorough discussion of
preprocessing technicalities and defer the reader to
Section~\ref{sec:experiments}. Equipped with this we may define what a
publication corpus is in this work.

\begin{definition}[Publication Corpus]\label{def:confdataset}
  A \emph{publication corpus} is a ternary relation $\mathcal{D} \subseteq P
  \times O \times Y$ where $P$ is a set of papers, $O$ is a set of publication
  venues (outlets) and $Y\subset \mathbb{N}$ is a set of natural number
  indicating the publication years. Furthermore, it holds $\forall (p,o,y)\in
  \mathcal{D}:|\{(o,y)\in O\times Y\mid(p,o,y)\in \mathcal{D}\}|=1$,
  i.e., there is a unique venue and year for each paper.
\end{definition}

We may note that \emph{papers, venues,} and \emph{year} are just
names. Hence, a publication corpus can be substituted by any structure
bearing the same construction, e.g., news articles in certain
newspapers on certain days.

\subsection{Document Representation in Topic Space}\label{sec:docrepr}

Say the number of papers (documents) in a publication corpus is $d$
and the set of all words from all these documents is of size $n$.
There are various methods for embedding these documents into a
topic space (commonly referred to as \emph{topic models}).  Most of
them employ a real-valued \emph{word-document} matrix as starting
point. Hence, we need to represent all elements $p\in P$ as elements
of $\mathbb{R}_{\geq0}^{n}$. Therefore we consider in the following
$p$ to be an element of $\mathbb{R}_{\geq0}^{n}$. In this
representation each vector component denotes a word-weight. Such a
weight for a particular word can be, for example, the word-frequency
(\emph{term-frequency}) or a more sophisticated approach like
\emph{tf-idf}~\citep{Ramos1999}. Having the vector representations of
all publications we can construct the word-document matrix $V$ simply
by juxtaposing all those vectors. A \emph{topic model} then finds a
representation of that word-document matrix $V$ as a product of two or
more matrices. These factors represent particular structural
elements of the resulting topic space, foremost the topics and some
representation for every document in this space, i.e., as a linear
combination of \emph{topic vectors}. An obvious goal here is that the
number of topics $t$ is substantially smaller than $d$.

A very prominent topic model, called \emph{Latent Semantic Analysis}
(LSA)~\citep{deerwester90lsa}, is based on the singular value
decomposition of the word-document matrix. This method has been proven
to work well for various natural language processing tasks, e.g., in
\cite{steinberger2007lsa}, in \cite{pu2006short} or for
recommendations of scientific articles in \cite{lsarecommendations}.
Nonetheless, we decided against its application in our work. The
reason for this is that the results are difficult to comprehend or
explain by humans. This problem arises from the fact that topics in
LSA can contribute positively and negatively to the topical document
representation.  Another well-employed method is \emph{latent
  Dirichlet allocation} (LDA) from \cite{lda}. It is based on a
probabilistic model where documents are assumed to be generated from a
distribution over topics, which themselves are distributions over
words. While LDA generally achieves very good results, e.g., in text
classification applications, it is known to fall short for small
documents like paper abstracts. This has for example been shown
empirically on Twitter posts in \cite{Hong2010}. Since we do prefer
for our modeling a method that is able to cope with small documents we
discard using LDA.  The same holds for well-known investigated methods
based on \emph{doc2vec}~\citep{doc2vec}. In doc2vec documents are
embedded together with words in a real valued vector space of a chosen
dimensionality. This is done such that related words and documents are
mapped closely together and unrelated ones far from each other, with
respect to some metric in that vector space. This method, however,
does not directly result in a topic space representation of the
documents.  While nearby words of a document give some idea of its
topic, these words may be different for each document. In contrast to
this, we require that every document is represented by the same
consistent set of topics. This allows for a meaningful comparison of
documents in terms of their topics as well as the analysis of topic
dynamics.

\subsection{Non-Negative Matrix Factorization}
\added{Review 1.7}{We now explain the used topic model NMF in more detail. Readers already familiar with or not concerned with the mathematical details of this method are advised to skip this section.}
We employ non-negative matrix factorization (NMF) \citep{Lee99MF} for
discovering paper topics and embedding papers into the topic space
$\mathbb{R}^t$ for a chosen dimensionality $t \in \mathbb{N}_{>0}$. We
start with a word-document matrix
$V \in \mathbb{R}_{\geq0}^{n \times d}$, with $n$ being the number of
words and $d$ the number of documents. In this matrix, each column is
a document vector in which each component is a word weight. A word
weight can be, for example, the term frequency in the document or the
product of the term frequency and inverted document frequency
(tf-idf). NMF finds an approximate factorization $V \approx WH$. The
factorization is achieved through minimization of a \emph{distance}
$d(V, WH)$, typically achieved through multiplicative update rules.  As
a distance $d$ we utilize (as commonly applied) the Frobenius norm of
the matrix difference $V-WH$, i.e., $d(V, WH)\coloneqq ||V-WH||_F$.
The Frobenius norm $||A||_F$ of an $m \times n$ matrix $A$ with entries $a_{ij}$
is $\sum_{i=1}^{m}{\sum_{j=1}^{n}{|a_{ij}|^2}}$. In
the following we will address this distance also by the name
\emph{reconstruction error}. After the minimization the left factor
$W \in \mathbb{R}_{\geq 0}^{n \times t}$ contains in each column a
topic vector of word weights. The factor
$H \in \mathbb{R}_{\geq 0}^{t \times d}$ contains in each column a
representation of a document in topic space. Typically this
representation is of much lower dimensionality than the original
document representation, i.e., $t \ll n$.

The reason for the superior interpretability of topic models obtained by NMF is that all elements in $W$ and $H$
are constrained to be positive. Hence, all components of a topic are additive.
Additionally, in the NMF variant we use the topic vectors in $W$ as well as the
document vectors in $H$ are constrained to have $L_1$-norm $1$. Therefore, a
document $V_{*i}$ (denoting the i-th column of V) can be (approximately)
reconstructed as the weighted sum $\sum_{j=1}^t{H_{ji} W_{*j}}$ of the topic
vectors $t_j \coloneqq W_{*j}$ found through NMF. More precisely, any document is a
convex combination of the topic vectors. The components $H_{ji}$ are interpretable as the proportion of the topics $j \in \{1 \ldots t\}$ in
the $i$-th document. The components of a topic vector $t_j$ can
be interpreted as the relative contribution of words to the topic. If the topic
vectors $t_j$ are linearly independent, they form a
basis of the $t$-dimensional topic space. Hence, the vector $d_i \coloneqq H_{*i}$
contains the coordinates of the $i$-th document embedded into this topic space.


\subsection{Venue Representation and Trajectories in Topic
  Space}

Since the ultimate goal of this work is to analyze the dynamics of
publication venues in topic space, we need to find a (computable)
representation of such venues, based on the so-far discussed document
representations. Hence, we decided to represent publication venues
through the centroid of their document embeddings. Employing centroids
appeared to us as a natural modeling decision, since they represent
the average of paper vectors, and thus the main topical research focus
of a venue.
Formally, for an index set
$I \subseteq \{1,\dotsc, d\}$ let the set
$D_{I} \coloneqq \{d_i \mid i \in I \}$ be the representations of
documents obtained through NMF, i.e., a set of the columns from matrix
$H$. We then represent this set through their centroid $r_I$:
\[r_I \coloneqq \frac{1}{\vert D_I \vert }\sum_{i \in I}{d_i}\]
A meaningful set $I$ may index all documents from a specific venue
 from a specific year $y \in Y$ or from a specific venue in
a specific year. By abuse of notation we may refer to the index set of a
specific venue $o$ and year $y$ by $I(o,y)$. Using this we may use the
shorthand $r_{I(o,y)}$ for the centroid of a specific venue $o \in O$
in a specific year. We note that this centroid does exist if and only
if the combination of $y$ and $o$ is present in the publication
corpus.

\begin{definition}[Topic Space Trajectory (Venue)]
  For a publication corpus $\mathcal{D}\subseteq P\times O\times Y$ and a venue $o\in O$ we call

  \[\tau(o) := \{ (r_{I(o,y)}, y) \mid \exists p \in P: (p,o,y) \in
\mathcal{D}\}\]
the \emph{topic space trajectory (of venue $o$)}.   
  
\end{definition}
Hence, to obtain a trajectory of a venue, we calculate one centroid
for each conference year. The set $\tau(o)$ can easily be linearly
ordered using the second element of all pairs $(r_{I(o,y)}, y)$ and
will therefore be considered a linearly ordered set in the
following. This order in time justifies the name trajectory. The idea
of venue trajectory is obviously adaptable to other entities, such as
sets of venues or author trajectories in topic space.



\subsection{Calculating an Optimal Topic
  Number}\label{sec:topiccoherence}
The utility of the just defined idea for topic space trajectories for
analyzing a publication corpus depends on the properties of the
concrete topic space. Most important is the number of topics. On the
one hand this number has to be large enough to discriminate elements of
the topic space. At the same time we require this number to be small in
order to maintain a human interpretability.
A general procedure to select an optimal topic number $\hat{t}$, is to calculate a
topic model for different values of $t$ as well as the value of a measure that rates 
the quality of the resulting topic model. For the final model, one
then chooses the $t$ with the maximum quality. In research on topic models and natural language processing in general, 
an established family of quality measures is called \emph{coherence measures}. As the name suggests, the objective behind these 
is to evaluate the topical coherence of a set of words (which, for example, represent a topic), i.e., how strongly the words are interconnected semantically.

The concrete coherence measure we use is called \emph{the $C_V$ measure} and has been described in \cite{Rosner2014}. The authors of that work have shown that amongst a variety of coherence measures, $C_V$ has the highest correlation with topic interpretability ratings from human annotators. Hence, employing this measure coincides with our goal of selecting topics with good interpretability. We will now describe the measure in detail. Although we did not invent it,
our contribution here is a much more concise
explanation of how to calculate it. In the original work, $C_V$
is embedded into an abstract framework for 
coherence measures. The explanation is therefore scattered in
text and formulas across many pages. We also give an interpretation of
the measure, which has not been done in the original
work.

Generally speaking, the $C_V$ measure for a topic is determined from co-occurrence statistics of the top-$n$ terms (ranked by their weights). The overall coherence of a model is calculated as the arithmetic mean of the topic coherences. The computation of the measure for one topic is achieved as follows: For each of the top $n$ words $w$ in the topic, a so-called \emph{context vector} $\vec{v}_w$ is created. The components of this context vector in the $C_V$ measure are the normalized pointwise mutual information ($NPMI$) between $w$ and all the top terms in the topic (including $w$ itself). As an example, let $n \coloneqq 4$ and let the top n terms of a topic be the set $\{search, query, engine, user\}$. Then the context vector for the term \emph{search} is determined via

\[\vec{v}_{search} = \begin{pmatrix*}[l] NPMI(search, search)^\gamma \\ NPMI(search, query)^\gamma \\ NPMI(search, engine)^\gamma \\ NPMI(search, user)^\gamma \end{pmatrix*}.\]
The parameter $\gamma \in (0, \infty)$ is used to put more or less weight on higher values of the NPMI. The NPMI between two words $w$ and $v$ is calculated as follows:
\[NPMI(w, v) = \frac{PMI(w, v)} {-\log{p(w, v) + \varepsilon}} = \frac{\log{\frac{p(w, v) + \varepsilon}{p(w)p(v)}} } {-\log{p(w, v) + \varepsilon}}\]

In the formula above, $p(w)$ is the probability of the word $w$ and $p(w, v)$ is the probability of $w$ and $v$ occurring together. The value $\varepsilon$ is a small constant to avoid logarithms of zero. The probabilities are determined from the same corpus on which the topic model is trained. More specifically, a sliding window of size $sw$ and step size $1$ is put over each document. Each step of the window over all documents defines what we will call here a \emph{pseudo-document}. The probability of a word (or two words) is then calculated as the number of pseudo-documents in which the term occurs (or both terms occur) divided by the total number of pseudo-documents. More formally, let $D$ be the set of pseudo-documents, and each $d \in D$ a set of words $w$.  Then
$p(w) \coloneqq |\{d \in D \mid w \in d \}|/|D|$
and $p(w, v) \coloneqq |\{d \in D \mid \{w,v\} \subseteq d \}| / |D|$. 
Once all context vectors have been determined, the coherence measure $C_V$ can be calculated as follows:
$$ C_V = \frac{1}{n} \sum_{i=1}^{n} { s_{cos}(\vec{v}_{w_i}, \sum_{j=1}^{n}{\vec{v}_{w_j}})}  $$
The cosine similarity $s_{cos}(w, v)$ between two vectors $w$ and $v$
is calculated by $\frac{w \cdot v}{||w|| \cdot ||v||}$ and it equals the cosine of the angle $\theta$ between the two vectors. Since the angle depends only on the direction of the two vectors, not their length, we can write $s_{cos}(\vec{v}_{w_i}, \sum_{j=1}^{n}{\vec{v}_{w_j}}) = s_{cos}(\vec{v}_{w_i}, \frac{1}{n}\sum_{j=1}^{n}{\vec{v}_{w_j}})$. The $C_V$ measure can thus also be interpreted as the average cosine similarity between the context vectors and their centroid $\frac{1}{n} \sum_{j=1}^{n}{\vec{v}_{w_j}}$.

The $C_V$ measure has three parameters: The number of top terms $n$, the size of the sliding window $sw$ and the weight $\gamma$. We follow the recommendations and results from the paper \cite{Rosner2014} and the default values in gensim of $n=20$, $sw=110$ and $\gamma=1$. Note that with our paper abstracts a window of size $110$ often contains the whole document. Hence, it might be worth considering smaller window sizes in future research. 

\section{Case Study on Machine Learning Publications}\label{sec:experiments}
We conduct a case study for topic space trajectories applied to a
machine learning paper corpus. The decision for using research
articles from machine learning is motivated by the fact that we
ourselves are researchers in that field and thus aware of its
subfields and development. We claim that the choice of a more
specialized topic does neither limit the applicability of our approach
nor the usefulness of interpretation methods, which will be presented
in~\cref{sec:interpret}.

\subsection{Dataset \& Filtering}
\begin{figure}
\includegraphics[width=1.0\columnwidth,trim= 8 0 9 0,clip]{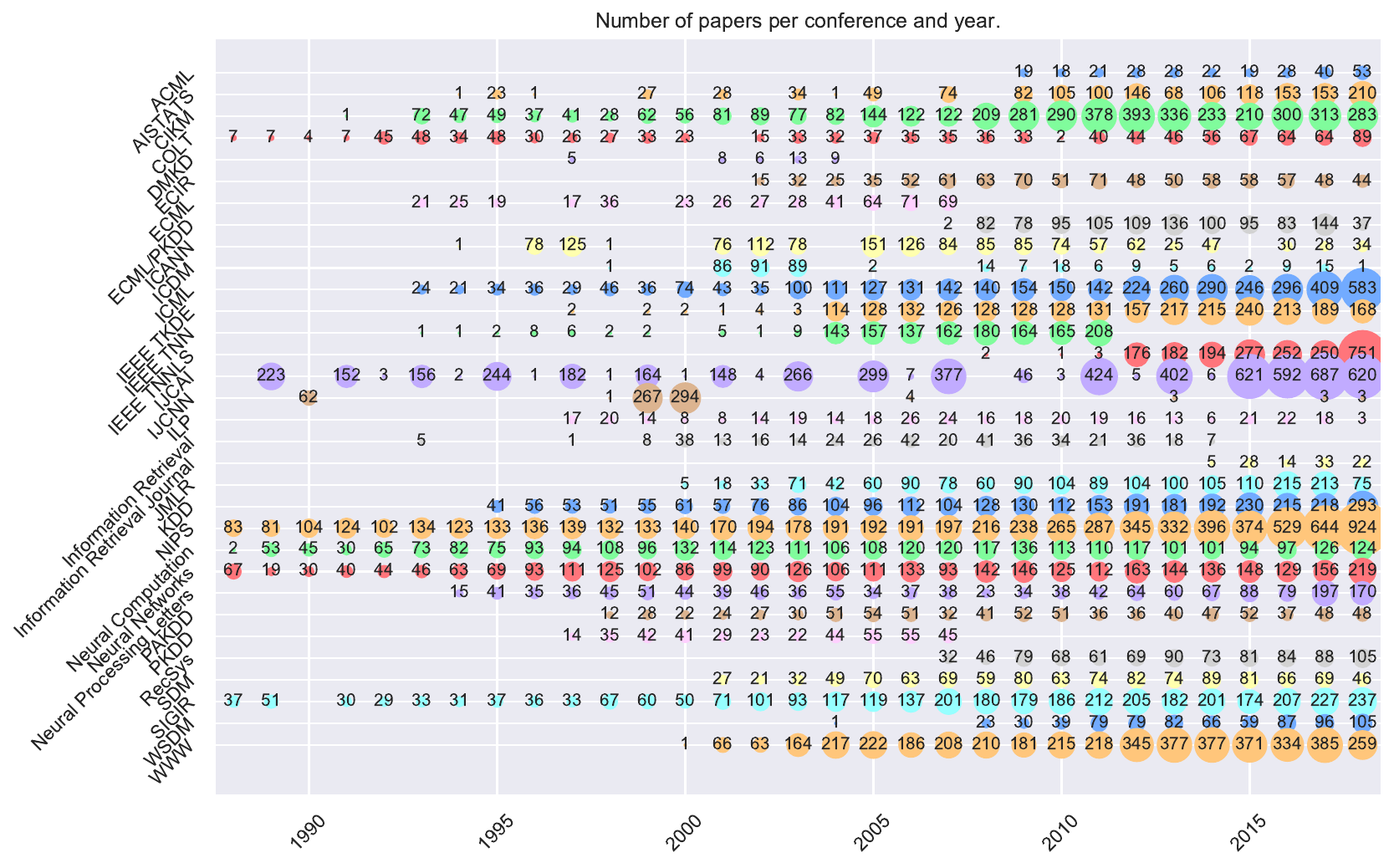}
\caption{Number of conference papers per year in our data set. The figure shows only a subset of the data. The area of each circle represents the number of papers in the conference.}
\label{fig:dataset}
\end{figure}
We use the Semantic Scholar Open Research Corpus for our analyses \citep{semanticscholar}. Our version of this corpus is the dump from January 31st, 2019. We extract papers from the top 32 machine learning conferences and journals compiled in \cite{kersting2019professur}. We also added the \emph{IJCAI} conference, which is another top-tier conference. For each paper, we have the paper title, the paper abstract as well as author and citation information. Hence, we emphasize here that our work is based on paper abstracts rather than full texts. Figure~\ref{fig:dataset} shows the number of papers in our data set by venue and year. When all papers are included, these date back until 1969. However, in this figure we only depict years, in which more papers from different venues are included. Since our dump of the Open Research Corpus is from early 2019, for this year it only contains a small number of publications. Hence, we also leave out 2019 in the figure.

To ensure a good coverage of venues, we manually compared the number
of papers of many randomly chosen years to the numbers counted in
conference proceedings and journal volumes. We estimate that for the
largest number of cases more than 80\% of published papers are covered
by our data. However, this is a very rough estimate and can sometimes
vary from venue to venue. For some of our analyses, we remove years or venues with very few papers from our data. In our estimation of conference topics, these few samples would otherwise lead to a skewed estimate of average topics. As an extreme example, we found a case, where the topic vector of a conference in a year was based on only one paper. In this case, the difference to previous and subsequent conference years was quite substantial, which made us aware of the problem.

We sometimes observed introductions to proceedings or journals being listed as separate publications in the data set. We remove these by heuristics based on the publication title. More specifically, we remove publications starting with the word ``Publication'' or including the substrings ``Introduction'' and \replaced{Review 2.9}{``special'' or ``Special''}{``pecial''}. We use this latter heuristic because the titles for some of such publications started with ``Introduction to Special Issue of'', where the word ``special'' is sometimes written in lower case and sometimes in upper case.  Although this simple detection mechanism could lead to false positives in some cases, we found it to work reasonably well on our subset of the Open Research Corpus.

A problem we found with the Open Research Corpus is that the abstracts
of many publications are missing. We experimented with representing
such papers only through their title. This, however, leads to
extremely short representations which do not cover well the actual
topical content. The main reason for this is their sparseness, i.e.,
few words being contained such that few co-occurrence statistics
between words can be gathered. Additionally, paper titles are
sometimes designed to be catchy instead of only representing content
or they contain the name of a self-developed method. Such titles can
be misleading for an automated analysis. Therefore we decided to fully
remove all papers without an abstract from our data set. 
\added{Review 2.5}{A different strategy here would be to incorporate other Scientometric data, such as citations or keywords or to merge in paper abstracts from further data sources. While it can be extremely useful to incorporate such further information, they either open up new problems on their own or, as in the case of keywords, can pose similar problems as paper titles. We hence restrict our work to incorporating only paper abstracts as a data source. Nonetheless, further data sources could open up interesting ways to the enhancements of our method in future research.} 
Altogether,
after this process, we are left with about 61 thousand machine
learning papers prepared for our analyses, i.e., which are equipped with a venue name, a title, a year and an abstract text and, additionally, the title of the work.

\subsection{Preprocessing}
For our analyses, we use the \emph{gensim} library for the Python
programming language \citep{vrehruvrek2010software}. This widely used
library provides implementations of many different topic models
together with various methods for natural language processing tasks,
e.g., preprocessing. For each paper, we only use the text from the
paper abstract. In \cite{scholz2014predictability} it has been shown
that paper abstracts are sufficient to represent the content of
publications and even can outperform models using the full paper
text. This is a natural consequence of the fact that abstracts consist
of a condensed summary of the research presented in the paper, but
without additional, more ``noisy'' discourses that might be provided
in the full text. We proceed by converting the texts to lower case,
tokenizing and removing stop words with the method
\emph{remove\_stopwords} from \emph{gensim}, which is based on a
\added{Review 1.4}{hard-coded list of stop words that apparently has been designed by the gensim developers.} Stop words are words like
\emph{``the''}, \emph{``such'}' or \emph{``and''}, i.e., words which
frequently occur in text but do not provide much topical
information. \added{Review 1.4}{The concrete list can be found in the gensim source code.}\footnote{\url{https://github.com/RaRe-Technologies/gensim/blob/develop/gensim/parsing/preprocessing.py}} \added{Review 1.3}{In natural language processing tasks, attempts are often made to reduce different word forms to a common word stem, e.g., through the Porter stemming algorithm. Stemming reduces the size of the vocabulary and thus leads to increased performance (in terms of speed) of subsequent calculations. However, it is also known to lead to word representations that are often no "real words" anymore. It is also known to reduce unrelated words to the same stem (e.g. \emph{university} and \emph{universe} both being mapped to \emph{univers}). Due to these disadvantages we do not use stemming.}
Finally, we represent each document in a vector space
using tf-idf weights. In this, we remove all terms from the model that
occur less than ten times in our data set. 
\added{Review 1.6}{Our aim here was to speed up subsequent calculations considerably without losing the very important words. The cut-off value is of course dependent on the corpus size. We determined its order of magnitude by looking at the term distribution that we will address in more detail in \cref{sec:datasetanalysis}. The reasoning here is that many terms exist, which are very unlikely to occur in a document and thus do not contribute much to calculated topics. Through their removal, we considerably reduce the size of the vocabulary and hence of the matrices in NMF. This improves memory usage as well as calculation times. A similar approach has been used in \cite{journals/ijufks/EckW07}, where concepts (small sets of words) appearing in less than ten documents were removed from analysis.}
Through this process, we
obtain the $n \times d$ word-document matrix $V=(w)_{ij}$ with entry
$w_{ij}$ containing the tf-idf-weight of word $i$ in document $j$.
The column vectors of $V$ equate the papers $P$, which together with other
data from the Open Research Corpus form a publication corpus
$\mathcal{D}$, cf~\cref{def:confdataset}.

\subsection{Matrix Factorization}
After having obtained a word-document-matrix $V$ as described in the
previous section, we apply to it the NMF-implementation from
\emph{gensim} using the default parameters. Since the minimization
process in this implementation of NMF relies on sampling as well as
randomly initialized matrices, it is nondeterministic. Hence, we
obtain different results in each run of the algorithm. Consistent to
the findings in \cite{topicstability}, our experience is that these
results are nonetheless very stable. This especially holds in
comparison to LDA, with which we also experimented. To determine the
topical stability, we manually compared the rankings of the top terms
of the learned topics. In almost all instances these ranked terms were nearly the same with slight variations in their order. Only some few topics were totally different in their content when running the learning algorithm repeatedly. We also briefly experimented with the parameters of NMF, namely the initialization method \added{Review 1.1}{(e.g., to a nonnegative double singular value decomposition instead of random values)} and the minimized reconstruction error, which can be changed to the Kullback-Leibler divergence. \added{Review 1.1}{Gensim does not have parameters to change the initialization method or the objective function of the minimization. We therefore experimented for this with the implementation from the python package \emph{sklearn}}. Subjectively we did not observe clear improvements or even strong differences and hence relied on the default parameters \added{Review 1.1}{of the gensim implementation}. \added{Review 1.2, 1.10}{To test for convergence of the gensim implementation of NMF, we plotted the error curves for repeated runs of the algorithm and with different parameters (such as a larger number of iterations over the data set). We found the error curve to very consistently converge to a low value after having seen about one third of our data set. Hence, no further parameter tweaking was necessary in our case.}

\subsection{Optimal Topic Number}
\begin{figure}[b] \centering
\includegraphics[width=0.50\columnwidth,trim= 13 8 21 10,clip]{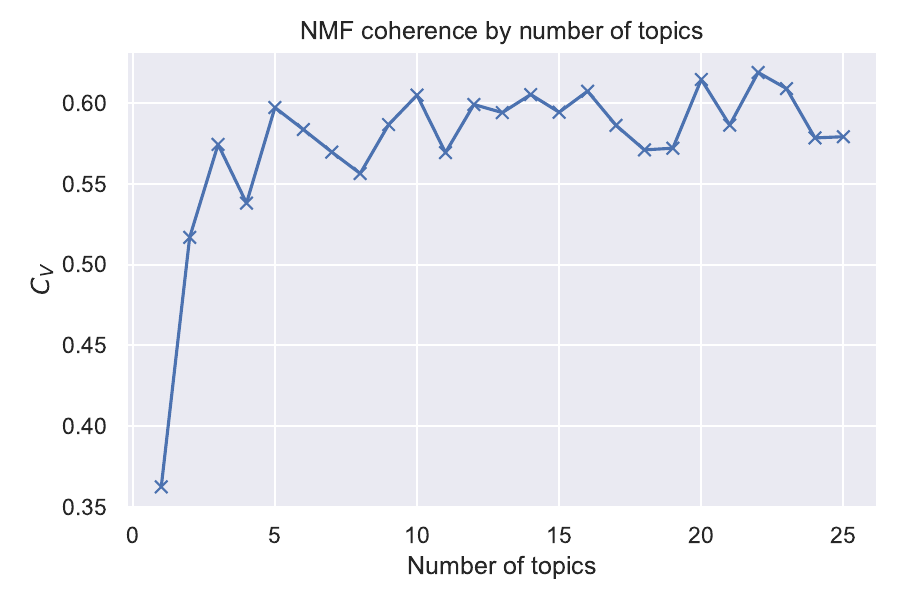}\hfill
\includegraphics[width=0.48\columnwidth,trim= 13 0 39 10,clip]{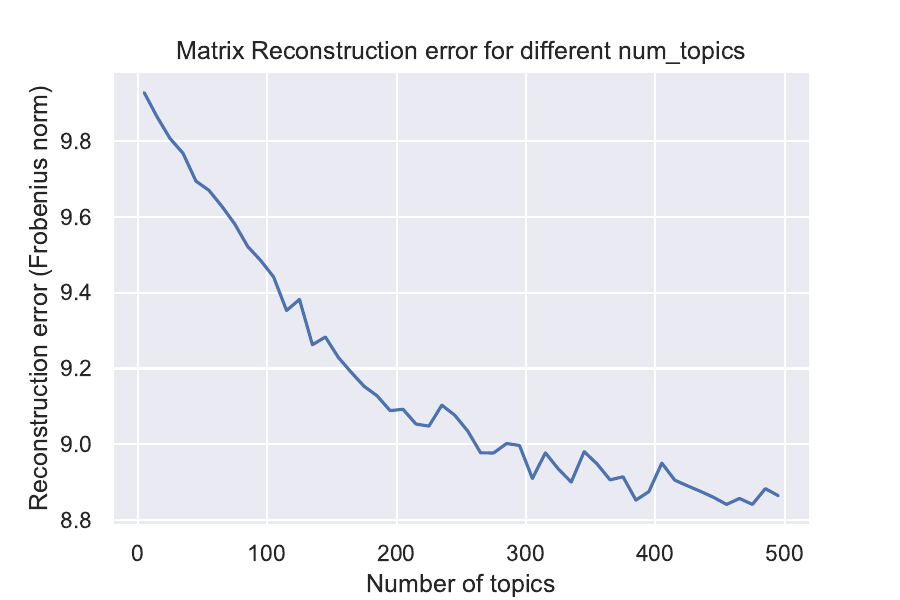}
\caption{We present for different topic numbers \textbf{Left:} The
  coherence measure $C_V$.\newline \textbf{Right:} The matrix reconstruction error.} \label{fig:topicnumbers}
\end{figure}

We calculated the coherence measure $C_V$ for every topic number from
1 to 25. The results of these calculations are given in
Figure~\ref{fig:topicnumbers} (left). We also calculated $C_V$ for some larger values, namely for each topic number $t$ from the set $\{30,
50, 100, 200, 300, 400, 500\}$. The results are not depicted here, since the coherences did not improve. We achieved the maximum coherence with
a topic number of $t=22$. We therefore use 22 topics in our subsequent
analysis. Note that for each topic number, we trained one single
instance of the NMF model. We did this with a fixed random seed to
ensure reproducibility. NMF is nondeterministic and the achieved model
quality could actually vary from run to run. Our reasoning here was
that NMF is still relatively stable. However, for future work we would
recommend to train multiple models for each topic number and use the
one with the maximum coherence. Using this method the full process of finding a model with a good topic number can be automated.

Before the $C_V$ measure, we experimented with other heuristics to select a topic number. As an example, we tried to take advantage of the matrix reconstruction error mentioned in section~\ref{sec:docrepr}. This error optimally should be a monotonically decreasing function of the number of topics, since the convex combination of more topic vectors should allow for an equally or more accurate document reconstruction. Our intuition was that plotting the reconstruction error for an increasing number of topics should result in a curve that starts with a strong (negative) slope and slowly converges to a minimal error with a slope close to zero. Our idea was then to select the topic number through the elbow criterion, i.e., by finding a point where the further decrease of the error with increasing $t$ is neglectible. Figure~\ref{fig:topicnumbers} (right) shows the curve resulting from this approach. While it matches our intuition, the optimal number of topics would lie somewhere near $500$. We find this number too large to give a comprehensible overview over research topics. We conjecture, however, that our approach would lead to success in selecting a good model for other downstream tasks such as classification.

\subsection{Abstract Length \& Term Distribution}\label{sec:datasetanalysis}
\begin{figure}
\includegraphics[width=0.5\columnwidth]{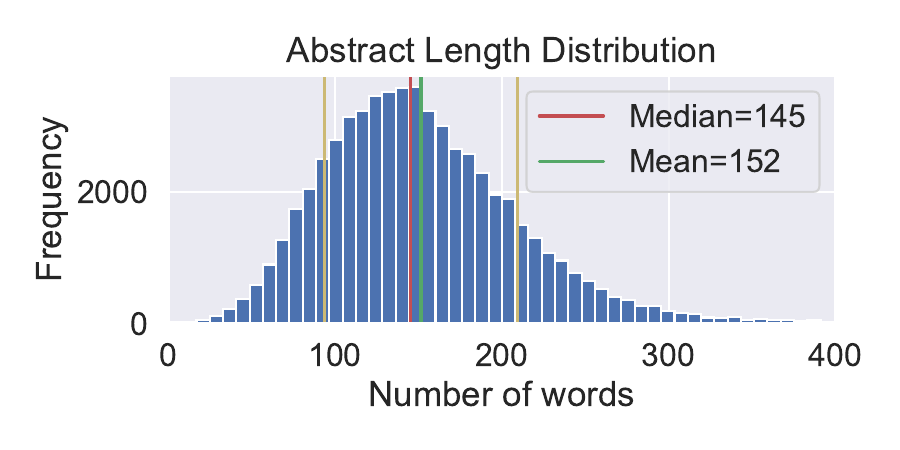}
\includegraphics[width=0.5\columnwidth]{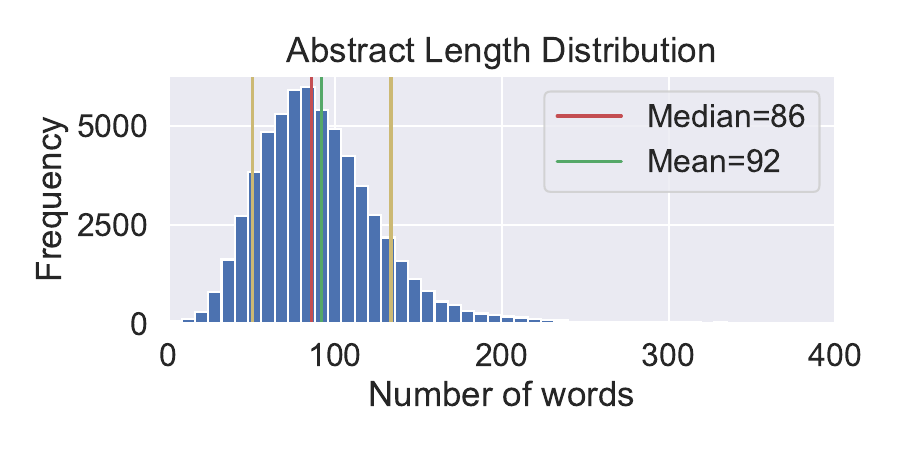}
\caption{Abstract length distribution. The figure shows histograms of abstract lengths (counted as the number of words) in our data set before (left) and after stop word removal (right). For this plot, we removed outliers, i.e., papers with extremely long abstracts, which occur due to wrong data in the corpus. Vertical lines give the mean, median, and standard deviation.} \label{fig:abstractlength}
\end{figure}
Since we work with relatively short paper abstracts instead of full paper texts, we deem it necessary to get an overview over some statistical properties of our data. Even more so, since our texts are shortened further through stop word removal and since real world data sets are sometimes erroneous, we need to ensure that mostly enough words are left with actual topical information. We also want to categorize our data in terms of a comparison to data sets from other text domains, to which our analysis methods could be applied.

First, we look at the distribution of the abstract lengths. In Figure~\ref{fig:abstractlength}, this distribution is given before and after stop word removal. We see that the average abstract consists of about 152 words, while the median is 145 due to a skewed distribution and some exceptionally long outliers. Paper abstracts are therefore longer than Twitter posts, which originally had a character limit of 140 that has nowadays been increased to 280. Hence, more information about the topic is included in a paper abstract. The order of magnitude, however, is still similar if we assume an average word length of five characters (a common assumption made in typing speed tests). Other document types such as web sites typically are several orders of magnitude larger in their word number. Once we remove stop words from the abstracts, their average length decreases quite substantially to only 92 words and a median of 86. Hence, about one third of words in a paper abstract are stop words with little information.

\begin{figure}
\includegraphics[width=0.5\columnwidth]{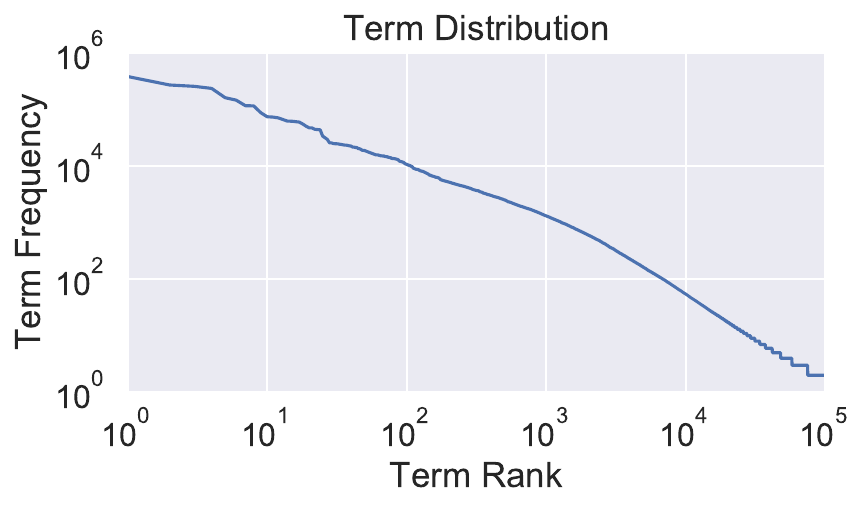}
\includegraphics[width=0.5\columnwidth]{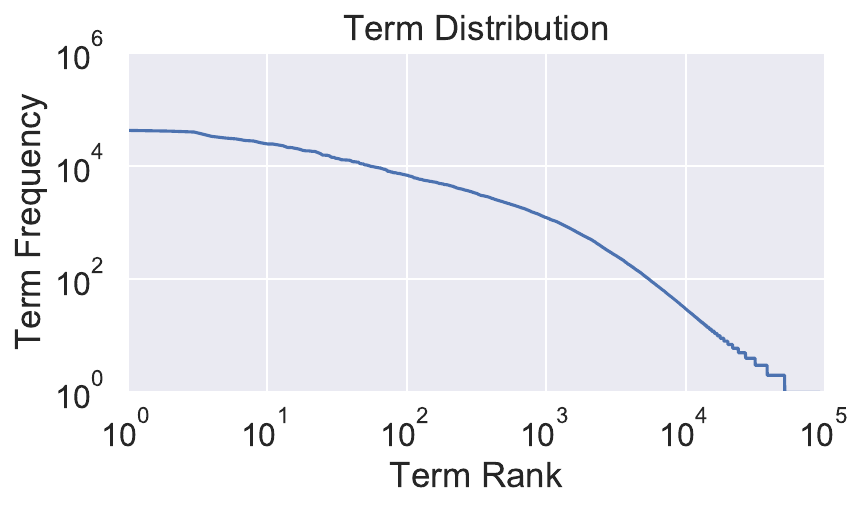}
\caption{Term distribution. The x-axis in both plots shows the rank of each term when sorted in descending order of term frequency. The y-axis shows the term frequencies. Term frequencies are given before (left) and after stop word removal (right).} \label{fig:termfrequency}
\end{figure}
Next, we analyze how frequently different words (terms)
occur. Figure~\ref{fig:termfrequency} shows the results of this
analysis in a log-log-plot before (left) and after stop word removal
(right). In the left plot we notice that before stop word removal,
term frequencies follow Zipf's law \citep{Zipf}, as expected. Zipf's
law is the empirical observation that when terms are ranked by their
frequency of occurrence, this frequency is distributed inversely
proportional to their rank, i.e., $P_n \sim 1/r^a$ for some
$a>0$. Stop words are words which occur frequently while containing
little information. Hence, after their removal, highly ranked words
are less frequent in the distribution as can be seen in
Figure~\ref{fig:termfrequency} (right). This leads to the removal of
words that are important for the matrix reconstruction error but have
no use for topic discrimination.


\section{Interpreting Topic Space Trajectories}
\label{sec:interpret}
After having described the circumstances of our case study, we are now
presenting the actual topic space trajectories that we calculated from
our paper corpus. What is more, we give interpretations and
visualizations of our findings. While we conducted our study on a
particular research field that we are familiar with, our approach here could be applied to other fields and text domains. Hence, our concrete analysis at this point simultaneously serves as an example of a generalizable approach. 

\subsection{Found Topics}

\begin{table}
\caption{Topics identified through NMF. Each topic is represented here by the top ranked terms. Terms are given in order of their rank (i.e., from highest to lowest word weight). Topics are numbered arbitrarily for reference. We manually assigned names to topics based on our own interpretation. Sometimes interpretations reflect a tendency or focus of a topic.} \label{tab:topics}


\scriptsize
\begin{tabular}{p{1.9cm} p{1.9cm} p{2.0cm} p{1.9cm} p{1.9cm}}
\toprule
\textbf{\mbox{1 Bayesian} \mbox{Inference}} & \textbf{\mbox{2 Search} \mbox{Engines}} & \textbf{\mbox{3 Neural} \mbox{Networks}} & \textbf{\mbox{4 Nonlinear} Control} & \textbf{\mbox{5 Optimization}}\\
\midrule
inference & search & network & control & convex\\
models & query & networks & controller & optimization\\
bayesian & queries & neural & adaptive & algorithm\\
model & engine & nodes & nonlinear & function\\
variational & engines & layer & systems & functions\\
latent & user & deep & tracking & convergence\\
probabilistic & click & training & feedback & gradient\\
variables & ranking & node & nn & linear\\
posterior & web & recurrent & robot & problems\\
gaussian & users & hidden & loop & algorithms\\
\vspace{-2mm}&&&&\\
\toprule
\textbf{\mbox{6 Neurons,} Dynamic Networks} & \textbf{\mbox{7 Classification}, Pattern \mbox{Mining}} & \textbf{\mbox{8 Information} Retrieval} & \textbf{\mbox{9 Social} \mbox{Media}} & \textbf{\mbox{10 Clustering}}\\
\midrule
neurons & tree & retrieval & social & clustering\\
neural & classifiers & topic & users & clusters\\
spike & classifier & topics & media & cluster\\
neuron & training & information & twitter & data\\
time & mining & relevance & content & algorithm\\
synaptic & decision & ir & influence & means\\
input & data & models & online & spectral\\
activity & classification & language & services & algorithms\\
spiking & trees & model & user & similarity\\
firing & ensemble & documents & people & sets\\
\vspace{-2mm}&&&&\\
\toprule
\textbf{11 Learning \& Knowledge Bases} & \textbf{\mbox{12 Semantic} Web} & \textbf{\mbox{13 Recommender} Systems} & \textbf{14 Graphs} & \textbf{\mbox{15 Reinforcement} Learning}\\
\midrule
learning & web & user & graph & policy\\
task & pages & recommendation & graphs & reinforcement\\
knowledge & page & users & nodes & agent\\
tasks & feature & items & edges & agents\\
entity & selection & item & subgraph & reward\\
entities & data & recommender & node & action\\
text & semantic & recommendations & structure & policies\\
fuzzy & content & collaborative & mining & value\\
online & information & filtering & vertices & learning\\
transfer & features & preferences & patterns & decision\\
\vspace{-2mm}&&&&\\
\toprule
\textbf{\mbox{16 Planning} \& \mbox{Reasoning,} \mbox{Association} Rules, Logic} & \textbf{17 Document Retrieval, STS-Tags} & \textbf{\mbox{18 Feature} Extraction, Dimension Reduction} & \textbf{\mbox{19 Support} Vector \mbox{Machines}} & \textbf{\mbox{20 Kernel} Methods}\\
\midrule
rules & document & feature & kernel & density\\
knowledge & documents & multi & svm & regression\\
rule & inline & features & kernels & estimation\\
logic & retrieval & classification & vector & kernel\\
language & query & class & stability & estimator\\
reasoning & math & dimensional & support & gaussian\\
planning & tex & sparse & matrix & distribution\\
semantic & formula & view & machines & estimators\\
semantics & term & supervised & svms & probability\\
domain & xml & dimensionality & delays & distributions\\
\vspace{-2mm}&&&&\\
\toprule
\textbf{\mbox{21 Matrix} Methods} & \textbf{\mbox{22 Image} Recognition}\\
\midrule
matrix & image\\
label & images\\
rank & visual\\
domain & object\\
labels & objects\\
learning & recognition\\
target & segmentation\\
labeled & features\\
data & scene\\
loss & spatial\\
\vspace{-2mm}&&&&\\

\end{tabular}
\end{table}

Table~\ref{tab:topics} shows the topics identified through NMF. Each
topic is represented by its top ten terms, determined from the word
weights in its topic vector. We manually assigned a number and a
name to each topic. The name is based on our own interpretation of the
top words in a topic. For this, we considered up to 50 terms per
topic. We may refer to some of these additional top words where we
consider it useful for the interpretation of a topic.

Most topics found through NMF are well interpretable and often clearly correspond to one specific research area. To name some, we found topics related to Bayesian inference, neural networks, nonlinear control, optimization, social media, clustering, semantic web, recommender systems, graphs, reinforcement learning and image recognition. However, we also found a few topics, which allow for ambiguous interpretations. As an example, topic 7 seems to be a mixture of pattern mining and classification methods. Topic 11 is a topic that appears to be related to different types of \emph{learning}, e.g. (based on more than the top ten words), \emph{transfer learning}, \emph{online learning}, \emph{representation learning}, \emph{reinforcement learning} and \emph{machine learning}. As such, it could be interpreted as a general topic on different concepts of machine learning, i.e., dealing with different approaches on \emph{how} and \emph{what} to learn. Additionally, this topic is related to knowledge bases. For some research areas, we found several topics with different focusses. As an example, we found two topics on neural networks (topic 3 and 6). Topic 3 here is more concerned with the architecture of networks (e.g., containing the words \emph{layers}, \emph{architecture}, \emph{structure}, \emph{convolutional}, \emph{feedforward}). Topic 6 puts more emphasis on neurons and the biological motivation of neural networks (e.g., \emph{neuron}, \emph{firing}, \emph{stimulus}, \emph{synapses}, \emph{brain}, \emph{signal}, \emph{cortical}). Additionally, topic 6 puts more emphasis on advanced, dynamic neural networks (\emph{spiking}, \emph{temporal}, \emph{recurrent}, \emph{memory}).

Topic 17 is a mixture of document retrieval and XML-Tags from the Standards Tag Suite (STS), e.g., \emph{tex}, \emph{math}, \emph{formula}. STS is an XML format used by publishers to exchange documents. In our data, STS was used for publications from one venue. Optimally, only the content from these tags should be parsed and added to the document representation. We spared this effort since few documents are concerned and a complicated checkup of the format of all documents with a subsequent parsing process would be required. Topic 18 is a mixture of feature extraction and methods related to dimensionality reduction (further top terms are \emph{manifold}, \emph{subspace}, \emph{embedding}). We consider this a sensible mixture since these two topics are strongly related, i.e., dimensionality reduction methods are often used to extract features from data.

Altogether we found that NMF gives topics with good interpretability. We also found some limitations of the method. In some cases, two different topics are mixed together, although they are not strongly related semantically, e.g., in topic 7 \emph{pattern mining} and \emph{classification}. We surmise this behavior of NMF is encouraged when some third terms often co-occur with both topics, e.g., here \emph{algorithm} and \emph{data mining}. It could especially be encouraged through polysemic or homonymous terms (i.e., terms that, in a different context, have slightly or totally different meanings). 
Second, NMF sometimes learns two topics that could be one, e.g., two on neural networks. This behavior tends to occur for topics that are overrepresented in the training corpus. Where desired, sub- or supersampling based on (research) categories and paper numbers of venues could therefore mitigate such results. A third limitation of NMF is its flat structure. NMF hence fails to convey the taxonomy of topics, e.g., search engines being a subtopic of information retrieval. However, this lack of complexity is at the same time an advantage, since it improves the comprehensibility.





\subsection{Topic Similarities}
\begin{figure}
\includegraphics[width=\columnwidth]{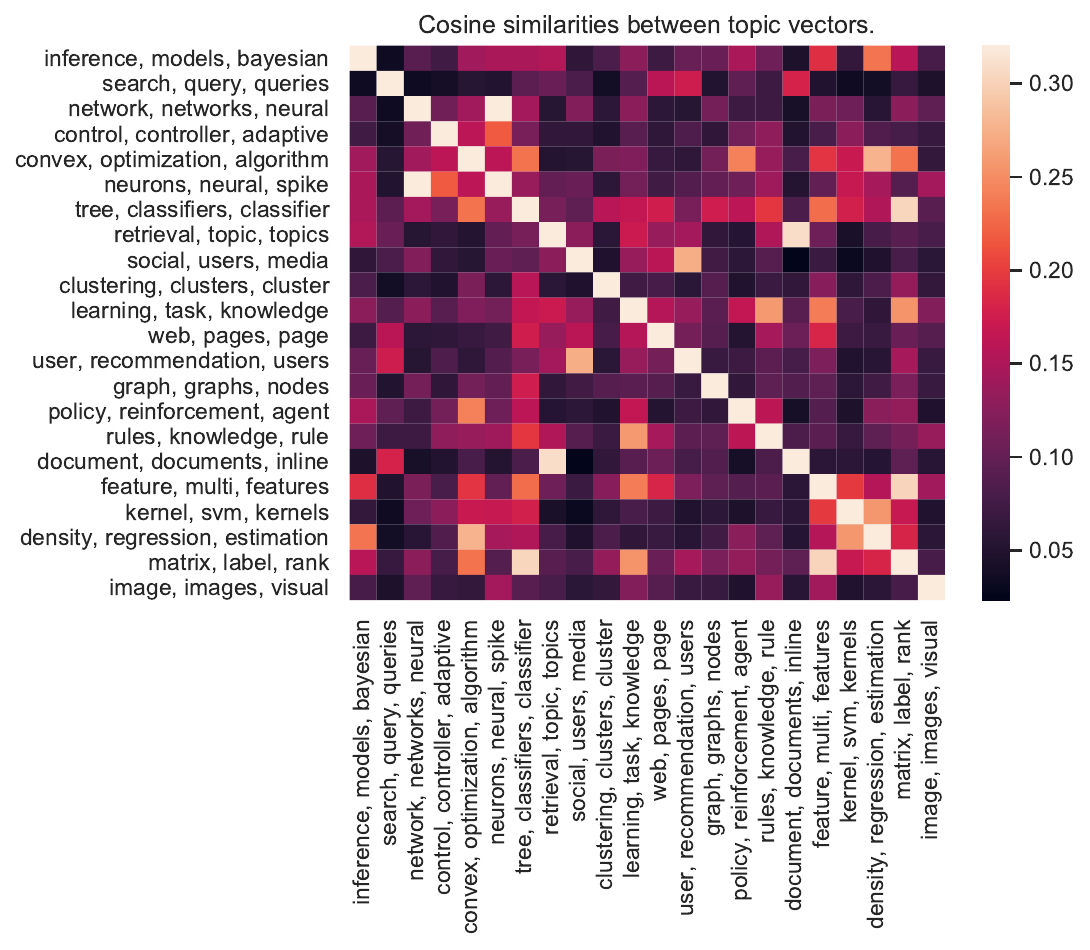}
\caption{Cosine similarities $s_{cos}(x,y)$ between all pairs of topic vectors $x$ and $y$. Note that similarities are symmetric, i.e., for all $x,y$ we have $s_{cos}(x,y)=s_{cos}(y,x)$. Diagonal elements have a cosine similarity of 1. Best viewed in color.}\label{fig:topicsimilarities}
\end{figure}

In figure~\ref{fig:topicsimilarities} we depict the cosine similarities between the calculated topic vectors $t_j$ with $j \in \{1 \ldots t\}$. Calculating the cosine similarity is a method often used in information retrieval to compare word weights of document vectors. The cosine similarity measures the cosine of the angle between two vectors. For vectors with non-negative components it lies between zero and one. Its maximum is reached for an angle of zero, i.e., when both vectors point into the same direction. Its minimum is reached here when the vectors are orthogonal. From the plot we can see how topics are related to each other. More specifically, brighter cells show that two topics have more similar word weights. Because the original topics have a dimension of more than 14000, it would be laborious to analyze this through a direct comparison of the word weights.

In the plot we see that most topic pairs have a low similarity. This indicates that most learned topics are a strong feature by their own and hence should enable us to represent well the variety of our data in topic space without redundancies. 
In the plot we notice that two topics on neural networks are most strongly related (\emph{``network, networks''} and \emph{``neurons, neural''}), but still have a low cosine similarity of about 0.3. We further see that two topics on information retrieval are closely related (\emph{``retrieval, topics''} and \emph{``document, documents''}). Such observations show that topics which come from the same research field (or supercategory) lead to more similar topic vectors. Turning this argument upside down, we can to some degree confirm or rebut our interpretations of the topic vectors. This is because similar topic vectors indicate similar research fields. It is interesting to notice that in some cases topics bear a comparatively high similarity to a variety of different topics. As an example, the topic \emph{``tree, classifiers''} is similar to almost each of the other topics. The topic \emph{``convex, optimization''} is similar to such topics as \emph{``tree, classifiers''}, \emph{``policy, reinforcement''}, \emph{``feature, multi''}, \emph{``density, regression''}, \emph{``kernel, svm''}. Supposedly, such cases occur due to a strong co-occurrence with other research topics and sometimes, these co-occurrences are also related semantically. As an example, optimization methods are often applied to a loss function for the training of classifiers such as some tree-based methods and support vector machines (SVMs), for fitting regression models, and in reinforcement learning. Hence, also the related topics occur together. The same holds for the topic on Bayesian inference, which is applied to a plethora of machine learning problems.

The lesson learned here is that research topics can be similar for at least three reasons: 1) They have the same supercategory. 2) One is a subcategory of the other. 3) One field is often applied to the other. Hence, calculating topic similarities enables us to interpret topics semantically to some degree. Note that the three cases can sometimes be hard to distinguish since learned topics sometimes do not reflect a pure, single research field. Methods for an automated analysis of these relations would open some interesting further research.

\subsection{Overall Historical Interest in Topics}

\begin{figure}
\includegraphics[width=\columnwidth]{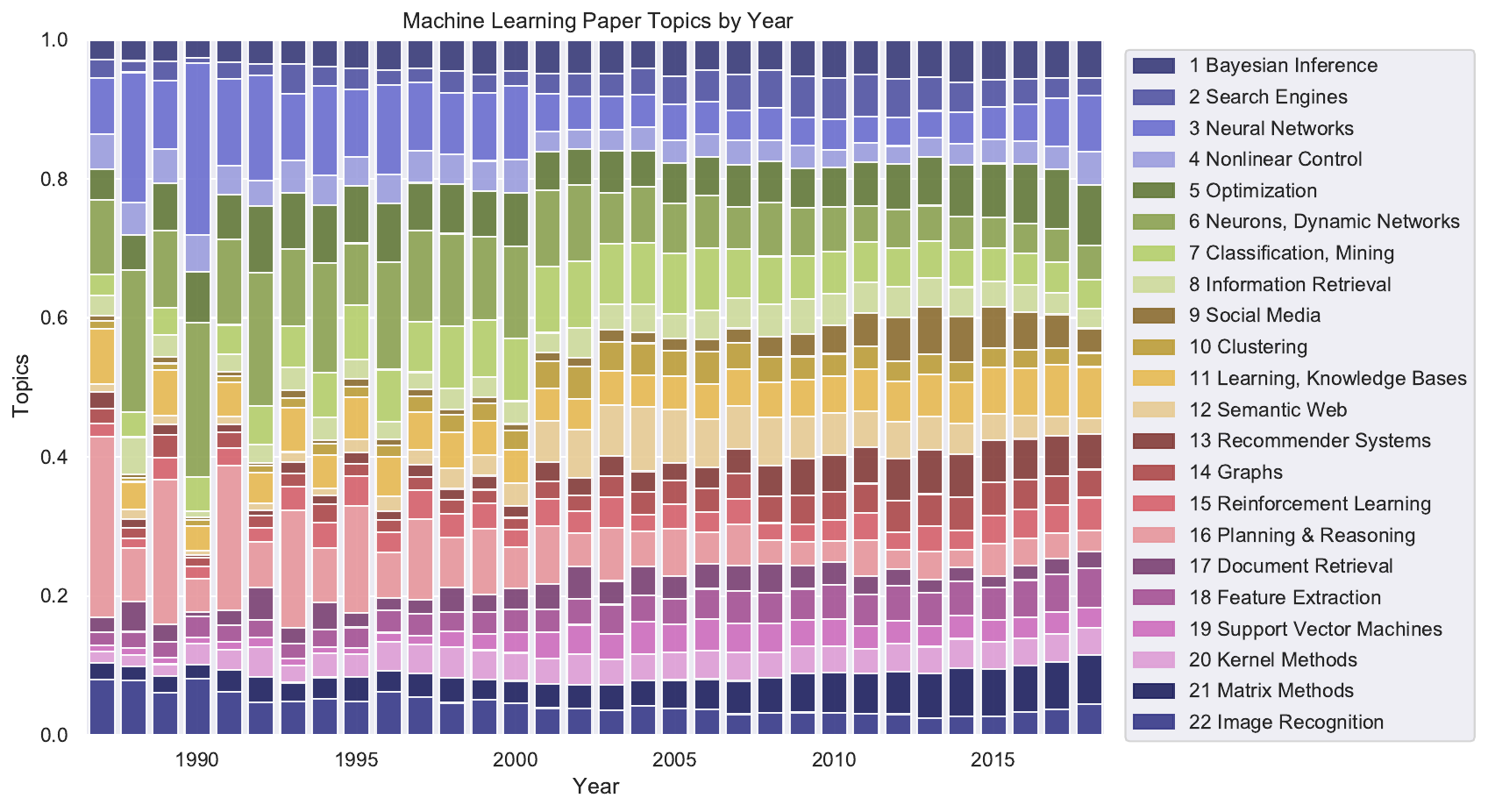}
\includegraphics[width=\columnwidth]{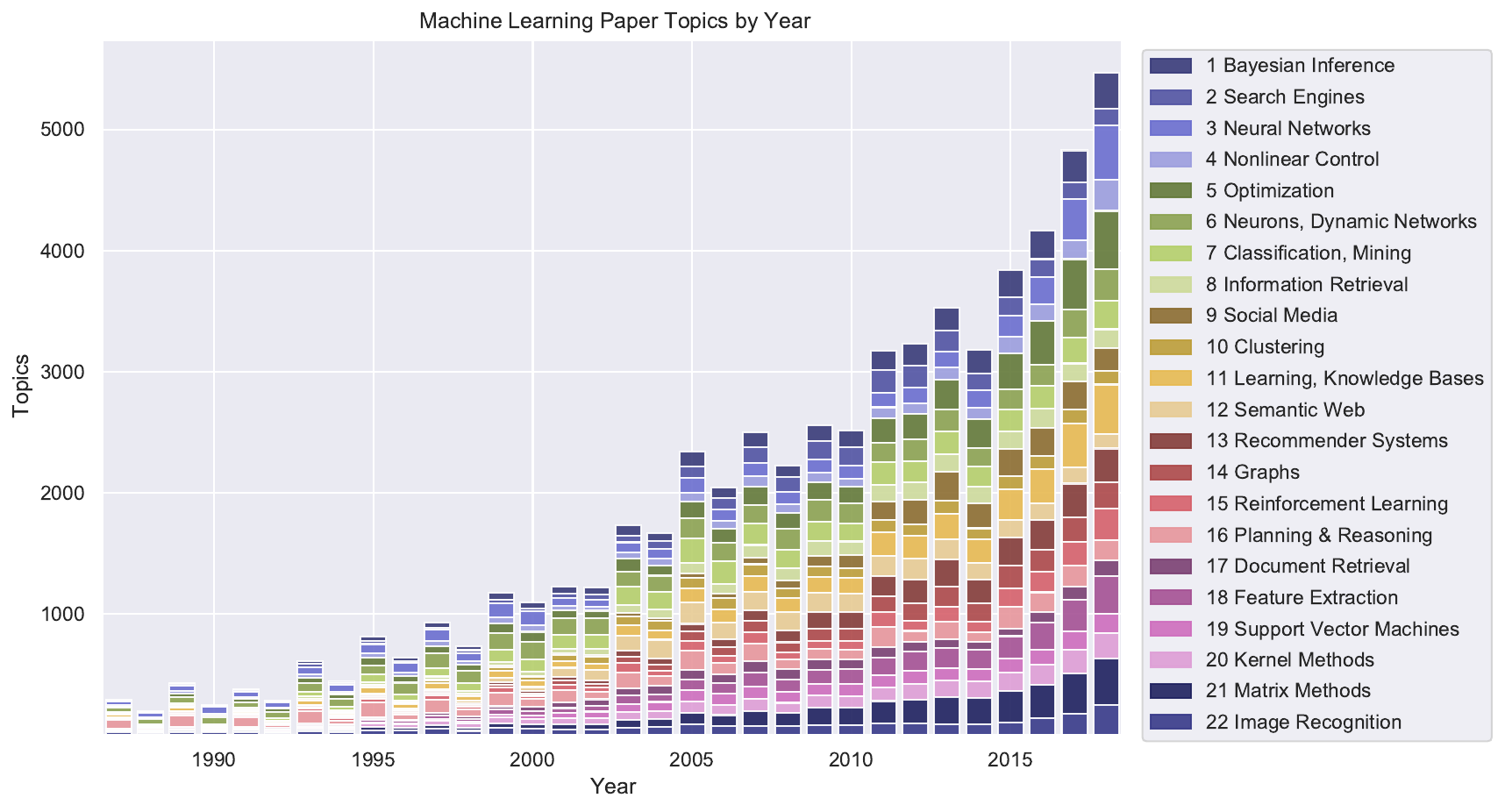}

\caption{Topic importances by years. Columns are chronologically sorted by years. For each year, the relative importance of each topic is given by the proportion of the bar height. In the top image, the stacked bars are normalized to height 1. In the bottom plot, the total height of stacked bars depicts the number of papers in our data set. Best viewed in color.} \label{fig:topicshares}
\end{figure}

In Figure~\ref{fig:topicshares} (top), we illustrate the topic trajectory for all papers in our data set from the years 1987 up to 2018. For this, we calculated one centroid per publication year. We depict the trajectory as chronologically sorted stacked bars, where each column depicts the topic space representation of one year. In Figure~\ref{fig:topicshares} (bottom), we calculated the sum of the document vectors without dividing by the publication number, as done for the centroids. Through this method, the topic weights are given proportional to the total number of papers in a year instead of relative to a total sum of 1. \added{Review 1.9}{Note that topic names and numbers correspond to the topics in \cref{tab:topics}. We sometimes used shortened names to improve the overview in our visualizations and will repeat on doing so in the remainder of this paper. However, we emphasize here that topics might comprise further notions than visible from the short names alone.}

We now analyze some striking results. For this, we refer to topics \replaced{Review 1.9}{}{by their top two terms and} by our manually assigned topic names.. Observing the topics from 1987 up to 2000 and comparing these with the years from 2001 to 2018, it becomes apparent that three topics with the initially largest weights lose their relative importance over the years. \replaced{Review 1.9}{These topics are topic 16 (\emph{Planning \& Reasoning}), topic 3 (\emph{Neural Networks}) and topic 6 (\emph{Neurons \& Dynamic Neural Networks}).}{These topics are \emph{rules, knowledge} (i.e., \emph{Planning \& Reasoning, Logic, Association Rules}), \emph{network, networks} (\emph{Neural Networks}) and \emph{neurons, neural} (\emph{Neurons \& Dynamic Neural Networks}).} Note that the bars here are influenced by the number of venues, that published papers in a specific year. This is why biennially some bars become larger and smaller. The IJCAI conference here was held every second year in uneven years, i.e., those where aforementioned topics have more weight. We also have venues that did not yet exist at the beginning of our analysis. In 2001, for example, the Journal on Machine Learning Research (JMLR) was introduced (cf. Figure~\ref{fig:dataset}). 2001 is also the first year in which a considerable amount of publications from the WWW conference appears in our data set (although founded earlier). Both these facts led to a larger number of publications in their specific research areas. Despite these influences, the overall tendencies of decline and raise in topic weights are visible. The mentioned decline of interest in neural networks took place at a time, when support vector machines (SVMs) became popular as a more efficient alternative (starting around mid-90s). SVMs stayed a widely used machine learning method. Neural networks, however, gained more interest again in more recent years. These facts become visible from our visualization and coincide with our personal background knowledge.

Further topics that have largely gained in interest include \emph{Social Media}, \emph{Recommender Systems}, \emph{Optimization}, \emph{Matrix Methods} and \emph{Bayesian Inference}, among others. Social media became more popular through platforms like Facebook or Twitter. The interest in recommender systems has been promoted through the \emph{Netflix prize}, a public contest on recommender systems that started in 2005, and through the introduction of the \emph{RecSys} conference in 2007. Besides this, the growth of online platforms such as \emph{Amazon} promoted the interest. Optimization, Bayesian inference and matrix methods have proven useful techniques that can be applied to a plethora of machine learning approaches. Topics that have recently lost weight in research are \emph{Semantic Web}, \emph{Search Engines} as well as \emph{Clustering} and \emph{Classification \& Pattern Mining}. Again, this coincides with our personal intuition. These latter two, very general topics have already been explored deeply in research and were often replaced by more specific problems, applications and methods.


Altogether, we notice that some machine learning topics gain, some lose relative importance. Sometimes, topics are almost invisible at the beginning and grow over time. This is a hint that these research topics just emerged or became more popular due to a certain event. Usually, popularity for machine learning methods is triggered by some milestone in their performance, i.e., when beating a benchmark on a data set by some large margin over previous methods. Sometimes other events trigger interest, such as public contests or the release of platforms.

\added{Review 1.11}{We also aggregated topics over all papers in our data set without differentiating by year. This gives us the overall prevalence of topics in our data set. What we found here was, that the most prevalent topics in our data set are \emph{Optimization}, \emph{Classification \& Pattern Mining} and the two topics on \emph{Neural Networks}. Each of these topics makes out about 6-7\% of our corpus. This is expected due to the many venues with large publication numbers on these topics in our data set. The prevalence of the least important topics, \emph{Document Retrieval/STS-Tags} and \emph{Clustering}, is only about half as high (about 3\%). This is partly due to their loss in popularity in recent years, where the overall publication numbers were higher.}


\subsection{Analyzing Venue Similarities through Topical Maps}\label{sec:confsims}

Topic space embeddings and trajectories are of a dimensionality that,
in general, cannot directly be depicted in a coordinate system. On the
other hand, it is often useful to visualize data in such a way. This
is because it allows to find similarities and differences between
entities (e.g., venues), indicated by nearness (or distance) in the
plot. To be able to visualize high-dimensional data in such a way, we
employ a well-known technique called multidimensional scaling (MDS), cf. \cite{mds}.
The idea behind MDS is to layout high-dimensional vectors in a low-dimensional space while preserving distances as best as possible. 
The low-dimensional representation of each input vector is found based on the squared differences between the pairwise distances of vectors in the input and the output space. More precisely, starting from random coordinates, points are aligned in the low-dimensional space s.t. an objective function is minimized as follows:
$$\min_{\hat{x}_1, \ldots, \hat{x}_n} \sum_{i = 1}^n {\sum_{j = 1}^n {\left(d(x_i, x_j) - ||\hat{x}_i - \hat{x}_j||_2\right)^2 }}  $$
In this, $n$ is the number of vectors, $x_i$ is one (high-dimensional)
input vector and $\hat{x_i}$ is the corresponding low-dimensional
output vector to be determined through MDS. The function $d$ is a
measure of distance in the input space. Here, we use the Euclidean
distance, i.e., $d(x,y) \coloneqq ||x-y||_2$. Note that in theory a
distance measure for compositional data such as the Aitchison distance
\citep{martin1998measures} or one between probability distributions,
e.g., those presented in \cite{wifidistances}, would be more suitable
for our data. In our investigation, however, we found that the Euclidean distance leads to a better separation of venues into three different research fields, namely \emph{neural networks}, \emph{information retrieval} and \emph{general machine learning}.

We use a dimensionality of two for the vectors $\hat{x}_i$, as this allows for good visualizations. In the resulting space, the two dimensions are not comprehensible as topics. Nonetheless, in this space we can analyze the topical similarity of venues based on their distance. 
\begin{figure}
  \begin{center}
    \includegraphics[width=0.8\columnwidth, trim={4mm 10mm 10mm 7mm}, clip]{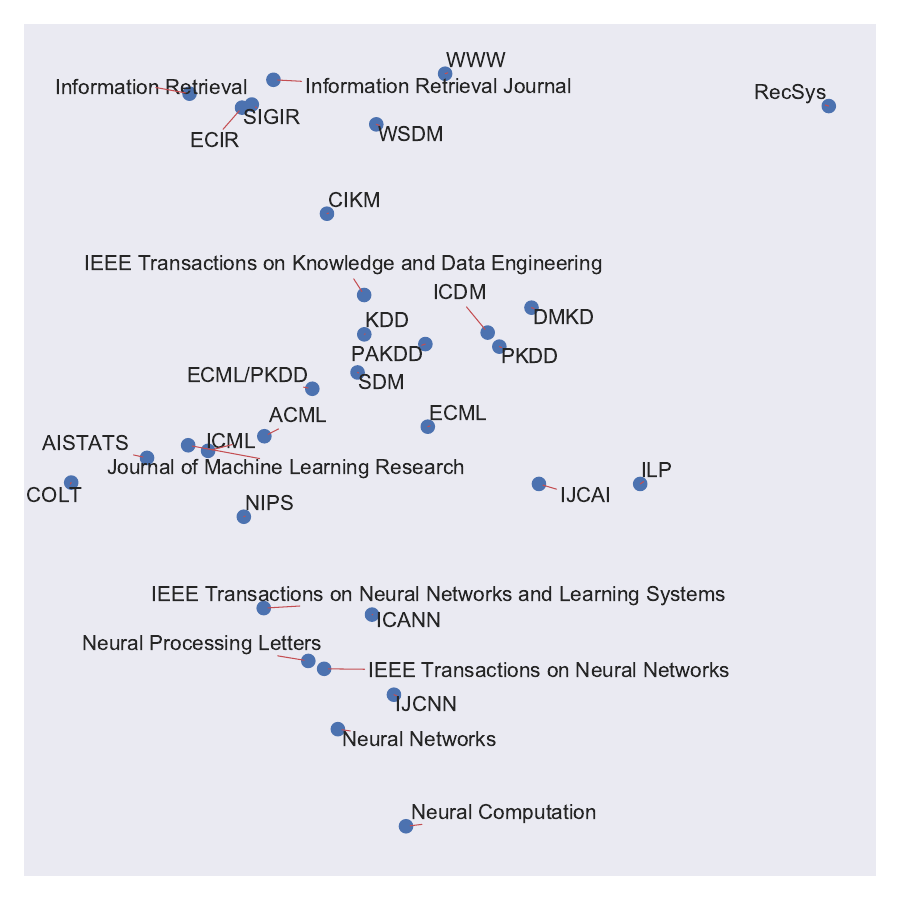}
  \end{center}
\caption{Topical map of venues. Dots show venue representations after dimensionality reduction through MDS. Closeness indicates topical similarity.}\label{fig:conferencemds}
\end{figure}
In Figure~\ref{fig:conferencemds} we depict topical representations of
venues, calculated as the centroids of all their papers' topic
vectors. We projected these centroids into a two-dimensional space
using MDS. In this figure, the closer points are together, the more
similar the research topics of their corresponding venues
are. Therefore we call a visualization like
in~\cref{fig:conferencemds} a \emph{topical map} of the venues from a publication corpus.

An interesting observation here is that venues cluster into three areas: In the top area we have venues that fall into the information retrieval category. In the middle part we have the more general machine learning venues. These two clusters are separated by \emph{CIKM} (\emph{Conference on Information and Knowledge Management}), which appears to fall somewhere in between both worlds. In the bottom area, we have conferences and journals specialized on artificial neural networks. \emph{RecSys} (\emph{Recommender Systems Conference}) in the top right of the plot is a conference with a strong topical focus on recommender systems. Hence, it appears to be a category by itself. However, RecSys lies most closely to the information retrieval world and is most dissimilar from neural networks.

Further interesting patterns emerge once we look at the specific venues. \emph{NIPS} (\emph{Neural Information Processing Systems}), as an example, was founded as a conference situated closer towards the neural networks topic. Over time, however, it developed into a more general machine learning conference, as can be seen by looking at current conference proceedings. Note that as a general tendency, the number of published papers per venue is growing from year to year as noticeable from Figures~\ref{fig:dataset} and \ref{fig:topicshares} (bottom). Hence, more recent publication years have a stronger influence on the centroid of a venue. This explains why NIPS falls into the general machine learning category. However, it clearly is the one conference from this category that is closest to the neural networks cluster. \emph{IJCAI} (\emph{International Joint Conference on Artificial Intelligence}) and \emph{ILP} (\emph{Inductive Logic Programming}, right middle) are both conferences which lie closest to the general machine learning category. Nonetheless, they put more emphasis on specialized topics, such as knowledge representations and logic based systems. Hence, they lie at the border to general machine learning with a small but visible gap to the central part. Similarly, \emph{COLT} (\emph{Conference on Learning Theory}) lies at the left border of the same cluster.  

In summary, the topical map shows that topics are captured well and as expected through our venue representations. Besides this, topical maps lead to interesting insights once we analyze (in this case visually perceived) clusters and edge cases, i.e., outliers and points which lie between several clusters. A natural enhancement of this method would be an analysis of \emph{trajectories} in topical maps, i.e., how venues drift apart or together over time.

\subsection{Visualizing Topic Space Trajectories through Projection}

\begin{figure}
\begin{center}
\includegraphics[width=0.7\columnwidth]{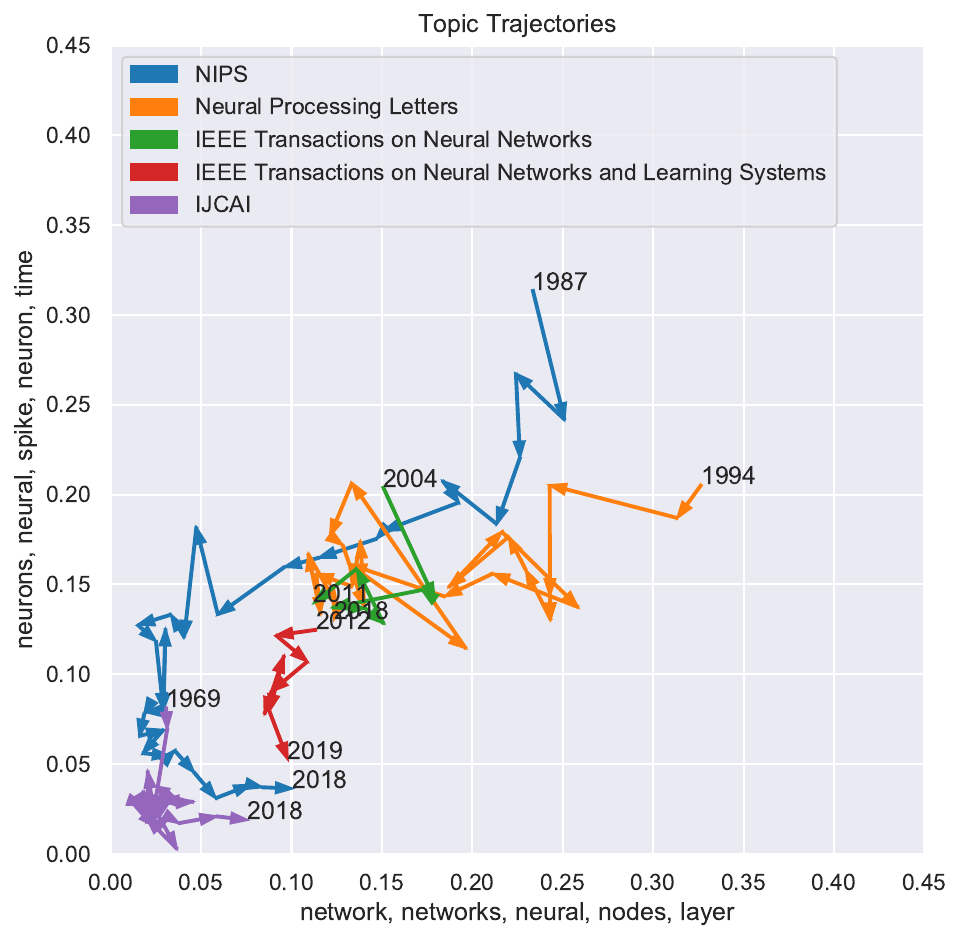}
\caption{Trajectories of selected Conferences. Projection on two selected topics on neural networks. Best viewed in color.}\label{fig:tps}
\end{center}
\end{figure}


Topic space trajectories exhibit too many dimensions (i.e., topics) for direct visualization in a coordinate system. To analyze trajectories, we hence project venue representations onto their most relevant two topics. We determine the relevance of a topic through its average weight in the trajectory, i.e., the topic weight averaged over all years. 
We demonstrate this method on the \emph{NIPS} conference, which we selected due to its interesting trajectory. We selected some additional venues, most of which are related to neural networks, for comparison. Figure~\ref{fig:tps} depicts the trajectories created through this process. We marked the first and last year of each trajectory. Trajectories drift into the direction of the arrows. Through measuring the average weight in the trajectory, we identified the two topics on neural networks as the most relevant ones. The topic on the x-axis is the one we previously identified as being more concerned with the architecture of neural networks (we simply called it \emph{Neural Networks}). The topic on the y-axis is more concerned with the biologic motivation of neural networks and with dynamic neural networks (i.e., recurrent and spiking neural networks). We called this one \emph{Neurons \& Dynamic Neural Networks}. 

We observe that initially all venues drift to the origin of the coordinate system, i.e., lose their relative interest in both topics. In recent years, however, research on neural network architecture has gained attraction again. By backtracking the trajectory of NIPS, we can see that this started around 2010. Around these years, the field of \emph{deep learning} \citep{journals/ftml/Bengio09} gained much interest. This interest was motivated and accelerated through several discoveries and improvements in the field, e.g., the ReLU (rectified linear unit) activation function in \cite{relu}, largely improved training times through GPU programming \citep{gputraining} and breakthroughs in performance on benchmark data sets, e.g., on the MNIST data set of handwritten digits in \cite{mnistbreakthrough} and on the ImageNet data set through convolutional neural networks in \cite{imagenetbreakthrough}. At this point we would like to note that other topics that gained interest, such as optimization, reinforcement learning and image recognition (cf. Figure~\ref{fig:topicshares}) are strongly connected to neural networks. Hence, the total interest in topics involving neural networks has increased even more than apparent from the trajectory in Figure~\ref{fig:tps}.

The conference IJCAI, which was held every two years starting from 1969 and yearly starting from 2015, is almost stationary most of the time. It exhibits only a small proportion of papers on neural networks. In recent years it shows a movement to the right, i.e., an increasing relevance of research on neural network architecture. The NIPS conference has a comparatively smooth, easy to follow trajectory, which again ends close to the origin. This result is supported by the fact that NIPS has become a more general conference on machine learning. \emph{Neural Processing Letters} starts close to NIPS but ends at a different location, with more relevance on both neural network topics. This is a reasonable result, since it is a journal focussed specifically on this research area.
An interesting case is \emph{IEEE Transactions on Neural Networks}, which has been renamed to \emph{IEEE Transactions on Neural Networks and Learning Systems} in the year 2012. In our data set these two are handled as separate venues. We noticed the name change through the behavior of the trajectories. More specifically, the endpoint of the trajectory under the first venue name lies close to the starting point of the trajectory under the second name.

Altogether, we see that topic space trajectories are an effective method for a human interpretable analysis of topic drift. A drawback resulting from high-dimensional data, like topic vectors, is that we can only visualize the trajectory for up to three topics. This problem, however, can be mitigated through the selection of relevant topics. One possibility here is a manual topic selection through the user. Likewise, an automated solution can be established through a measure of topic relevance. For our example in Figure~\ref{fig:tps}, this measure is the average weight of the topic. We can imagine further measures for different applications. Using the maximal topic weight over all years would yield trajectories for topics which were strongly relevant, even when only for a short time. Measures based on the absolute difference between topic weights for different years open another promising direction. Such measures would return trajectories with strong movement in topic space. This can be fine-tuned based on which years are considered (e.g., the difference between the first and the last year, between all consecutive years or between all pairs of years) and how these are aggregated (e.g., using the average or maximum of the differences).

\subsection{Visualizing Topic Space Trajectories as Heat Maps}

\begin{figure}
\includegraphics[width=\columnwidth]{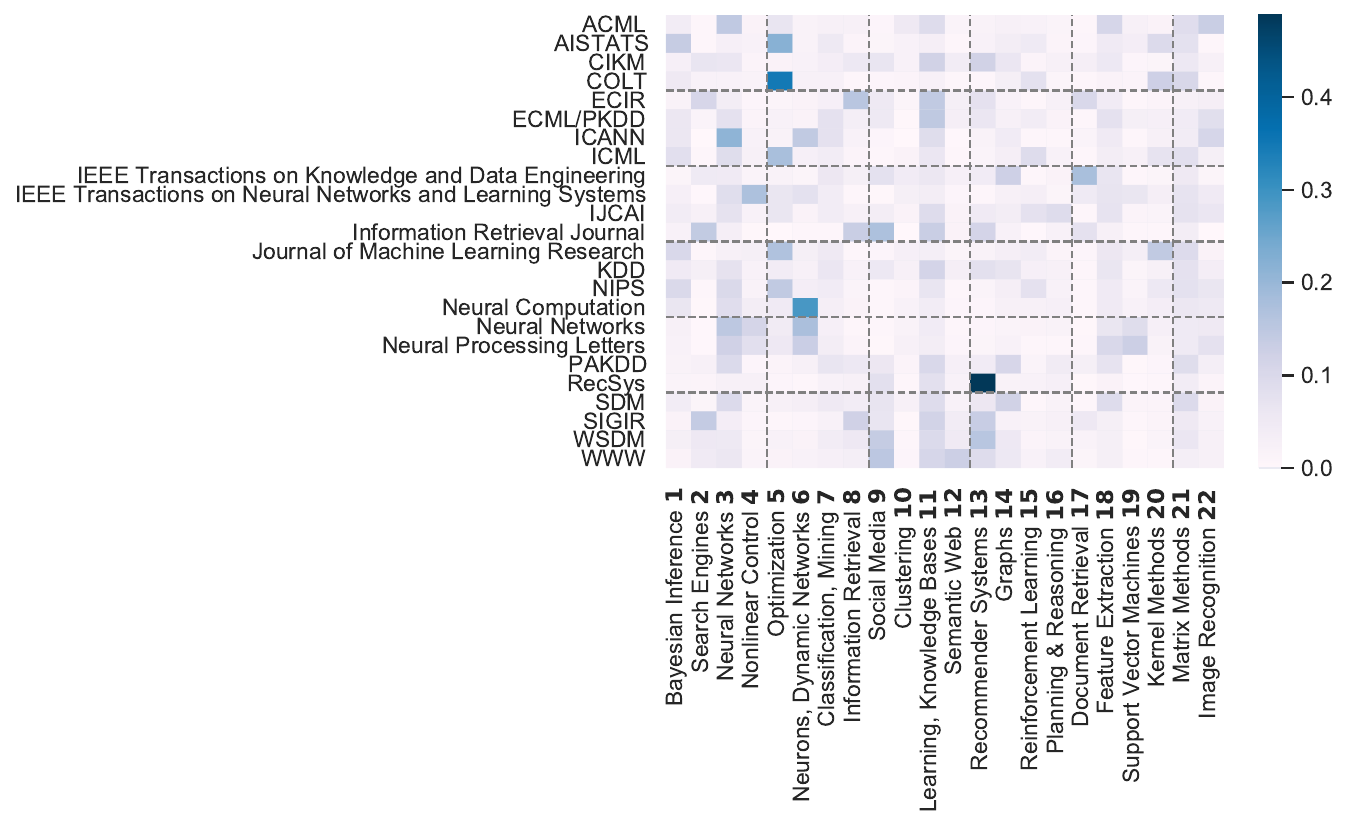}
\caption{Conference topics for 2018. In this visualization, each row represents a conference. The conference is represented through their mean topic vector. Each column represents a topic. Colors indicate topic weights. For each topic, the most important terms are given.} \label{fig:confheatmap}
\end{figure}

\begin{figure}
\includegraphics[width=\columnwidth, trim={2.9cm 2.7cm 2.9cm 7cm}, clip]{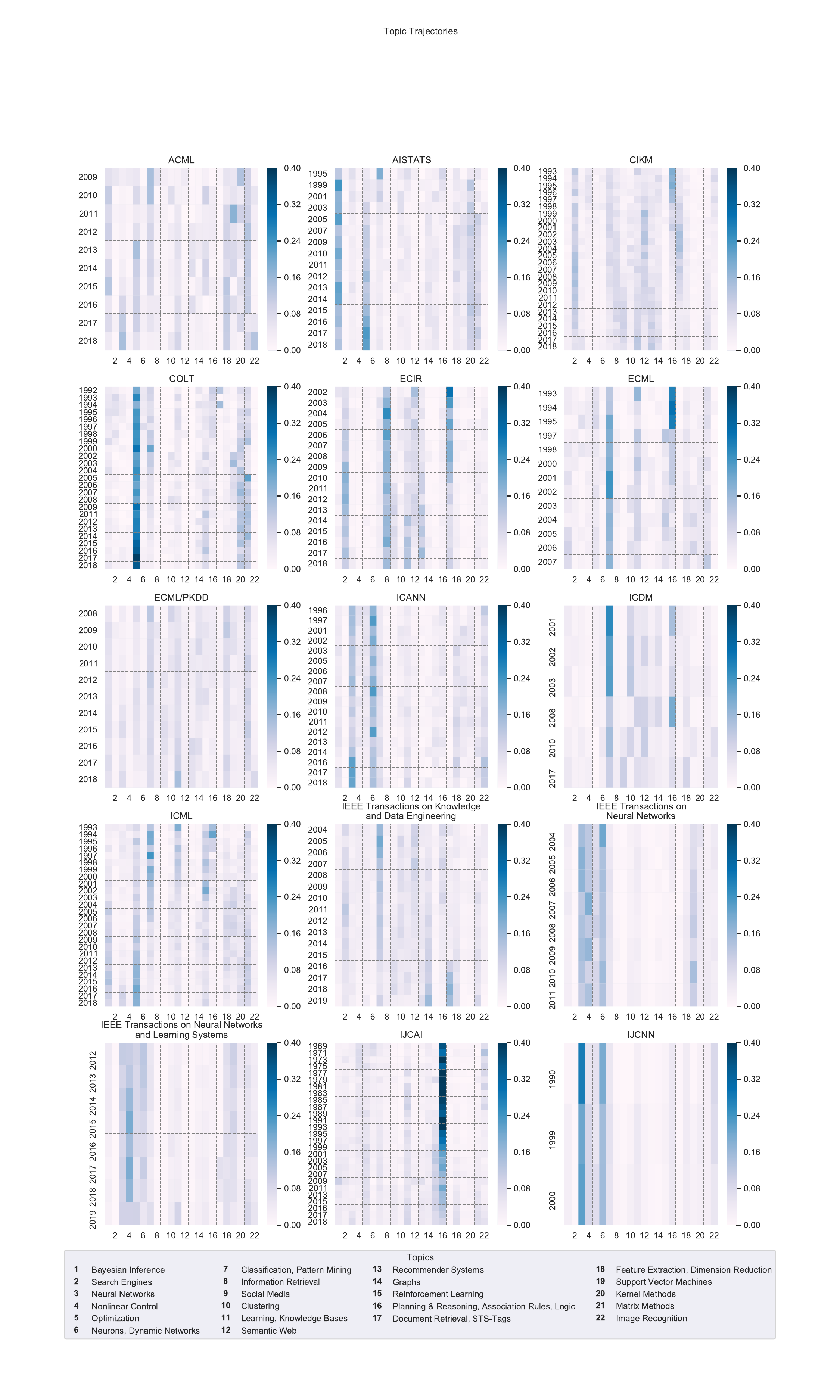} 
\caption{\textbf{(a)} Topic space trajectories as heat maps. Each heat map visualizes the topic space trajectory of a venue. In each heat map, rows represent years and columns represent topics. Cell colors indicate topic weights. \replaced{Review 1.9}{Topic numbers are given on the x-axis in each subplot and corresponding topic names in the legend at the bottom.}{For each topic, the most important terms are given.} DMKD is missing since we require a minimum number of ten papers per year. \added{Review 1.9}{Trajectories are continued on the next page.}} \label{fig:confheatmaps}
\end{figure}
\addtocounter{figure}{-1}

\begin{figure}
\includegraphics[width=\columnwidth, trim={2.9cm 8cm 2.9cm 8cm}, clip]{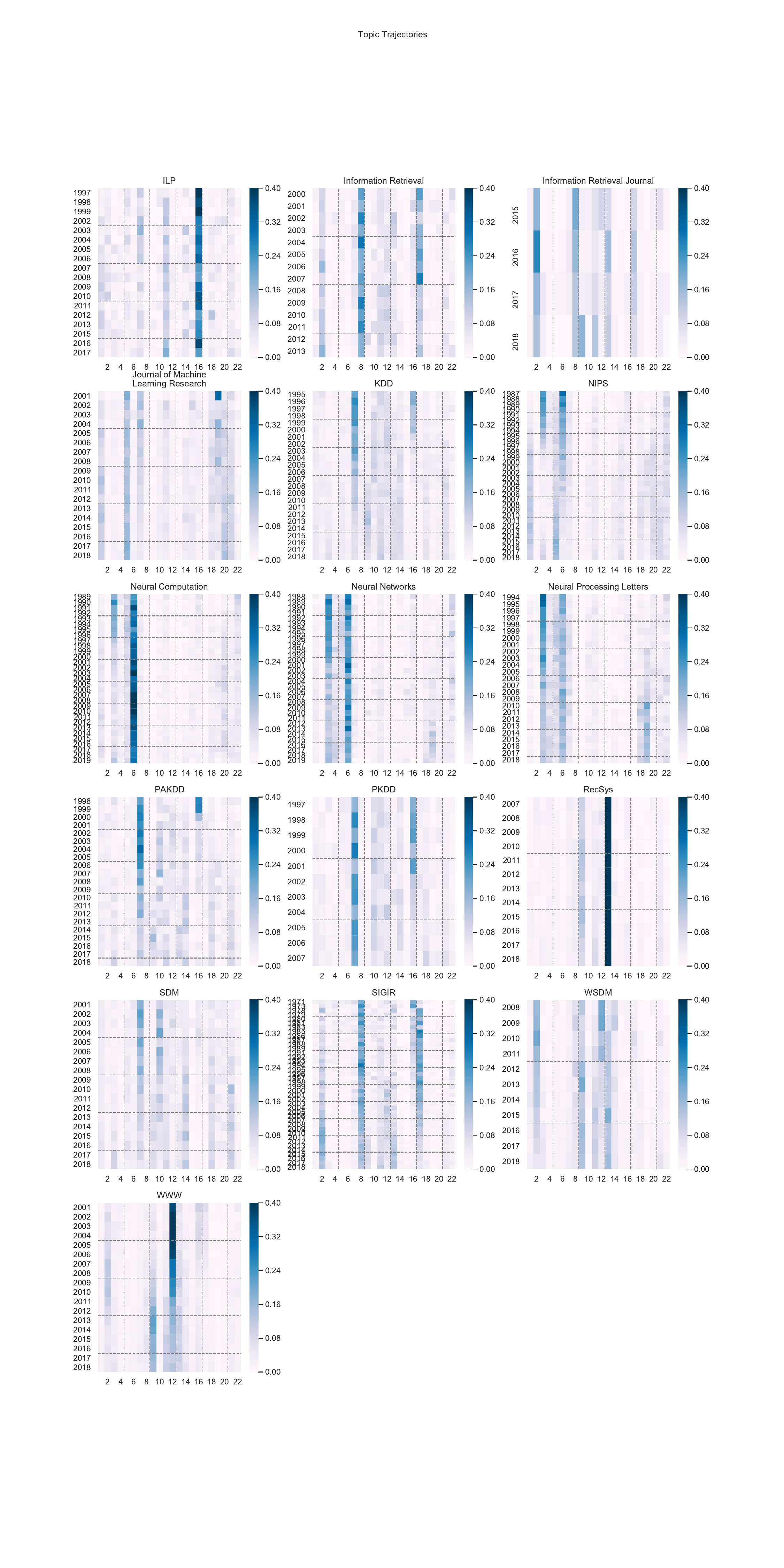} 
\caption{\textbf{(b)} \added{Review 1.9}{Topic space trajectories as heat maps (cont'd).}} \label{fig:confheatmaps2}
\end{figure}

We found that heat maps as depicted in Figure~\ref{fig:confheatmap} are an effective method for interpretable topic visualizations. In such plots, topic weights can effortlessly be compared across venues. Each row here visualizes the topic space representation of one venue. In this specific instance, we calculated the centroid of papers from 2018. The topic weights are visualized through different color shades in the columns, with brighter colors indicating stronger topic weights.

Building upon this idea, in Figure~\ref{fig:confheatmaps} \added{Review 1.9}{(a) and (b)} we visualize all topic space trajectories of our data set through one heat map per venue. In these heat maps, each row represents the topic space representation of a venue for a specific year. By following the rows from top to bottom, we see how interest in specific topics evolves over time. We only calculated and displayed centroids for years, in which at least ten papers were published at a venue. We do this for two reasons: First, occasionally we have instances of papers which seem to have a wrong year or venue. Second, sometimes our data set contains very few papers for a year. Both these lead to venue representations, that do not reflect reality well. We hence only calculate trajectories over years with more samples, i.e., papers. This leads to the effect, that we have no trajectory at all for the venue \emph{DMKD}.

In the resulting heat maps we make two particularly interesting general observations: First, the topics evolve smoothly from year to year, despite the fact that each row was calculated from completely different papers.
Second, we see that venues with similar research exhibit similar patterns in their heat maps. The heat maps thus can be regarded as distinguishable fingerprints that research areas leave. As an example, the heat maps of different venues on Information Retrieval (e.g., \emph{Information Retrieval} and \emph{SIGIR}) exhibit a visually similar appearance. This is due to their topic weights being stronger in the same columns as well as being similar in their development over time (i.e., over the rows). Likewise, such a similarity is strongly visible for conferences and journals on artificial neural networks (e.g., \emph{NIPS}, \emph{Neural Computation} and \emph{Neural Networks}). In particular, we have two important topics on neural networks with a similar development over time.


Altogether the strongly visible patterns show that topic space trajectories capture topical specifics about venues as well as about their historical development. Our visualizations as heat maps make these specifics visually perceivable. We argue that heat maps are among the best possible visualization method for these kinds of trajectories. The reason is that heat maps capture all dimensions, i.e., topics, while providing a good and comparable overview.

We now analyze the trajectories of some particular venues. One important machine learning conference is the \emph{International Joint Conference on Artificial Intelligence} (IJCAI). What strikes most for this venue, is that it has a strong focus on one particular topic, which is \emph{Planning \& Reasoning, Logic \& Association Rules}. This focus, however, has decreased since 1995 and the conference has become more diverse. In particular, \emph{Reinforcement Learning} has gained importance since then, which is a different approach to solve similar tasks.

Another interesting case are the \emph{European Conference on Machine Learning} (ECML) and \emph{Principles and Practice of Knowledge Discovery in Databases} (PKDD). Both conferences, similarly to IJCAI, started with a high interest in planning \& reasoning (etc.) that declined since 1995 (for PKDD some years later). Both additionally exhibit a strong focus on \emph{Classification \& Pattern Mining} throughout their existence until 2007, as well as some interest in \emph{Clustering}. What distinguishes these conferences is how weights are distributed across other topics. PKDD here is concerned with web pages, graphs and knowledge bases. ECML is concerned with matrix and kernel methods as well as SVMs and feature extraction. In 2008 these two conferences were merged and since then called \emph{ECML/PKDD}. The resulting heat map of this conference thus exihbits even more diversely distributed topics than the two alone.

\subsection{Topic Diversity}
\begin{table}
\caption{Ranking of venues by topical diversity. Topical diversity is measured as the effective number of species, which is the exponential of the Shannon entropy. In this table, horizontal rules are displayed after every ten ranks to improve readability. \added{Review 2.7}{Journals are printed in italic. There is a tendency of journals to be more focussed.}} \label{tab:confentropy}
\begin{tabular}{r | l r}
\toprule
Rank & \replaced{Review 2.7}{Venue (Conference or \emph{Journal})}{Conference} & Diversity\\
\toprule
\textbf{1} & ECML/PKDD & 19.30\\
\textbf{2} & \emph{IEEE Transactions on Knowledge and Data Engineering} & 19.24\\
\textbf{3} & KDD & 19.08\\
\textbf{4} & CIKM & 18.21\\
\textbf{5} & ACML & 18.02\\
\textbf{6} & SDM & 17.99\\
\textbf{7} & PAKDD & 17.54\\
\textbf{8} & NIPS & 17.13\\
\textbf{9} & ICML & 16.64\\
\textbf{10} & IJCAI & 16.51\\
\midrule
\textbf{11} & ECML & 16.36\\
\textbf{12} & ICANN & 16.25\\
\textbf{13} & \emph{Journal of Machine Learning Research} & 16.18\\
\textbf{14} & \emph{IEEE Transactions on Neural Networks and Learning Systems} & 16.01\\
\textbf{15} & ICDM & 15.81\\
\textbf{16} & PKDD & 15.44\\
\textbf{17} & WSDM & 15.42\\
\textbf{18} & AISTATS & 15.25\\
\textbf{19} & SIGIR & 14.58\\
\textbf{20} & ECIR & 14.42\\
\midrule
\textbf{21} & \emph{Neural Processing Letters} & 14.35\\
\textbf{22} & WWW & 14.15\\
\textbf{23} & \emph{IEEE Transactions on Neural Networks} & 14.10\\
\textbf{24} & \emph{Information Retrieval Journal} & 13.96\\
\textbf{25} & DMKD & 13.92\\
\textbf{26} & \emph{Neural Networks} & 13.74\\
\textbf{27} & IJCNN & 13.73\\
\textbf{28} & \emph{Information Retrieval} & 13.49\\
\textbf{29} & COLT & 12.95\\
\textbf{30} & ILP & 12.82\\
\midrule
\textbf{31} & \emph{Neural Computation} & 11.73\\
\textbf{32} & RecSys & 7.87\\
\bottomrule
\end{tabular}
\end{table}

In this part, we analyze the topical diversity of venues based on their topic space representations. For this, we calculate a measure of diversity from each of these vectors. As the components of the vectors are all positive and sum up to $1$, we can interpret each topic space representation as a probability distribution over topics. A higher diversity should be obtained, the more evenly distributed these topics are. This can be achieved through the Shannon-Entropy. The Shannon-Entropy of a probability distribution $p$ over a discrete random variable $x$ with outcomes $X$ is calculated as follows: 

\[
H(p) = -\sum_{x \in X}{p(x) \cdot \ln{p(x)}}
\]
While this measure becomes larger, the more evenly distributed topics are, its concrete value is not well interpretable. In \cite{jost2006entropy}, the entropy is therefore converted to a more interpretable measure through taking the exponential, i.e. by calculating $exp(H(p))$. This measure is often used in biology to calculate \emph{the effective number of species}, i.e., the number of evenly distributed species that would be necessary to obtain the same calculated entropy. Hence, the maximum possible value of this measure for a distribution over $n$ outcomes is exactly $n$. This is reached when all outcomes have the probability $1/n$. In information theory, the measure is also sometimes referred to as the \emph{perplexity} of a distribution, although with a different connotation and interpretation.

We calculate the effective number of species for the topic space representation of every venue. We then rank the conferences in decreasing order of diversity. The results are given in Table~\ref{tab:confentropy}. The results agree with our background knowledge about these venues. More general conferences or journals on knowledge discovery, which contain research from almost every machine learning field are ranked highly in the list. First rank is ECML/PKDD with a diversity of 19.30. This is closely followed by other more general conferences and journals on machine learning and knowledge discovery, such as KDD and CIKM. In contrast to this, in the lower part of the table we see venues which are more specialized. As an extreme example, RecSys, with its strong focus on recommender systems, is ranked lowest with a diversity of 7.87. Second lowest rank is Neural Computation with a diversity of 11.73. COLT and ILP are comparatively specialized conferences as well. Almost all other conferences from the lower part of the table (ranks 19-28), are specialized conferences or journals on neural networks or information retrieval.

In the middle part of the table we have the cases which are in between. Interesting examples are ECML and PKDD, which starting from 2008 merged to the single conference ECML/PKDD. While both already exhibited a strong topical diversity, it increased even more through the merge. Another interesting case is NIPS (Conference on Neural Information Processing Systems). Although this originally was a conference on neural networks as its name indicates, it later evolved into a more general conference on machine learning and artificial intelligence. Hence, its topical diversity is considerably higher than the diversity of all other conferences and journals focussing on neural networks.

\begin{figure}
\includegraphics[width=\columnwidth]{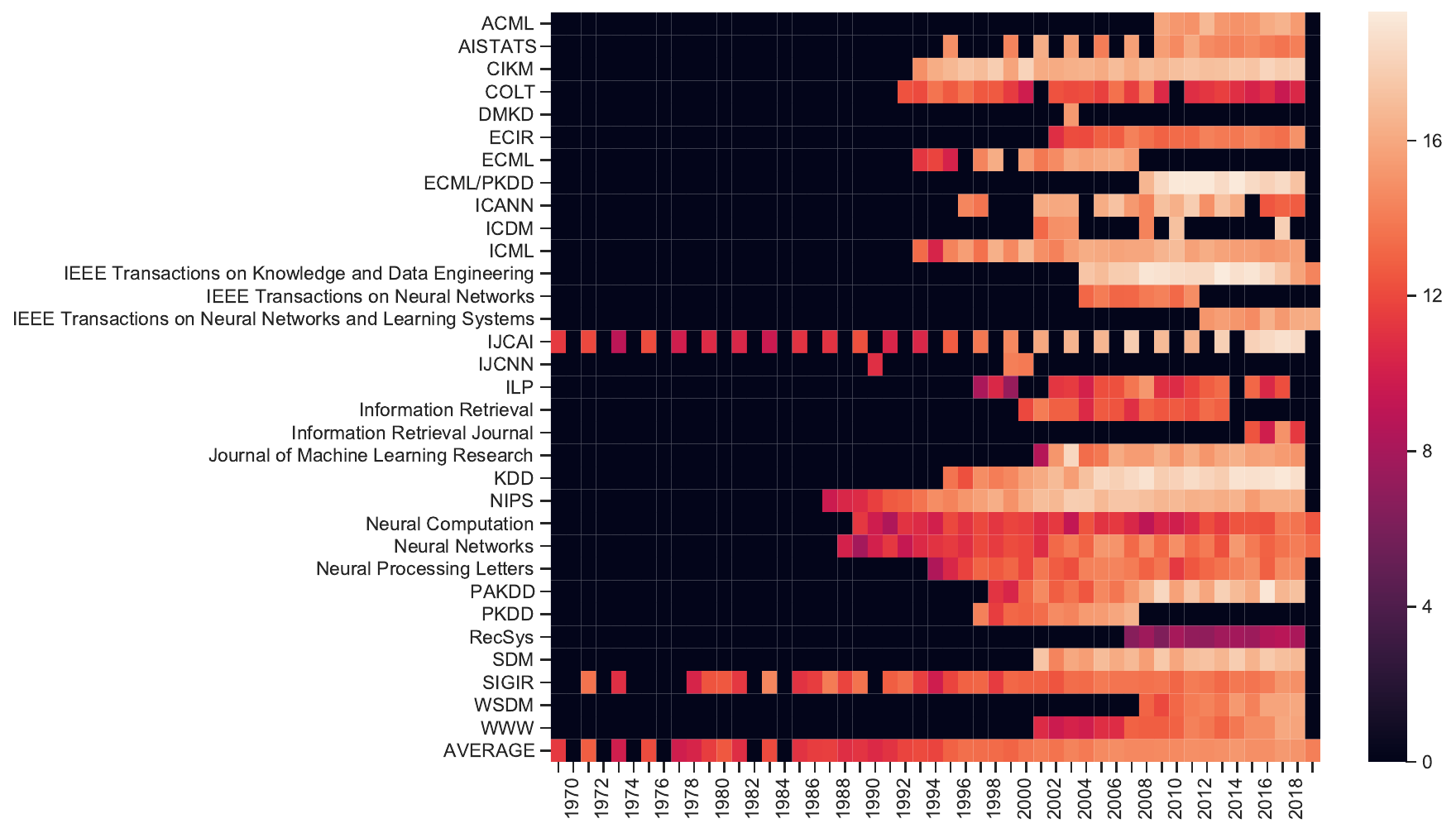}
\caption{Topic diversities for venues over years. The diversity of a venue in a specific year is indicated by the color (or shade) in the plot. Black boxes indicate that not enough data is available. In most cases this means, that the venue did not take place or exist during that year. The last row contains the average diversity over all venues in that year.}\label{fig:topicdiversity}
\end{figure}
In a second analysis, we look at the development of diversity over the years. This is depicted as a heat map in Figure~\ref{fig:topicdiversity}. In this figure, each row depicts the topic diversity of a venue for all years our data set spans. The diversity is indicated by the color (or shade) of a cell in a row. The last row contains the average diversity over all conferences. Diversities were only calculated, where at least ten papers were available for a venue and year. Note that in some cases, we still have erroneous data. As an example, SIGIR did not take place before 1978. Analyzing the results, it is interesting to note that many venues start out with an increase in diversity (e.g., NIPS, WWW, KDD). In later years, however, and especially the last few years, the diversity often is lower than previously (e.g., NIPS, ICML, IEEE Transactions on Knowledge and Data Engineering, AISTATS). The tendency of an early increase is also noticeable in the last row containing the average diversity. Here, however, we see that later on the diversity is close to constant. It seems that during the last years, some conferences put more focus on specific topics again, while on average the diversity does not change. One reason for this could lie in the growing number of different conferences and journals. Venues here sometimes might want to distinguish themselves more from others. This leads to more focussed single venues, while the average diversity remains nearly constant. Venues could also have become more selective in their review process, choosing only papers that fit the conference well. This could be a consequence of the increasing popularity of machine learning in the recent years, which has also led to a larger number of submitted papers that venues can choose from.

Another interesting case is the WWW conference. From 2001 to 2006, this conference has a diversity that ranges between 9.7 and 10.7. In 2007, there is a sudden increase to a diversity of 12.9. Our topic trajectory heat map in Figure~\ref{fig:confheatmaps} indicates that in this year the previously strongest topic on web pages has lost popularity. Instead, interest in search engines and social media has grown. Additionally, the topic on recommender systems starts to gain relevance, although already starting one year before in 2006. We assume that two events played a big role for this result: In 2006, the social media platform \emph{Twitter} was released. Twitter quickly became a popular resource for research due to its large, global user base and public API. The second event was a competition on recommender systems, called the \emph{Netflix prize}\footnote{\url{https://www.netflixprize.com}}. In this competition, which started in 2006, participants were invited to develop a recommender system that predicted user ratings for films. The prize money of one million dollars led to an increasing interest in recommender systems with more than 5000 teams actively participating in the competition. 
\subsection{Topic Densities}\label{sec:topicdensity}

\begin{figure}
\includegraphics[width=0.5\columnwidth, trim={1cm 0.5cm 1.5cm 0.5cm}, clip]{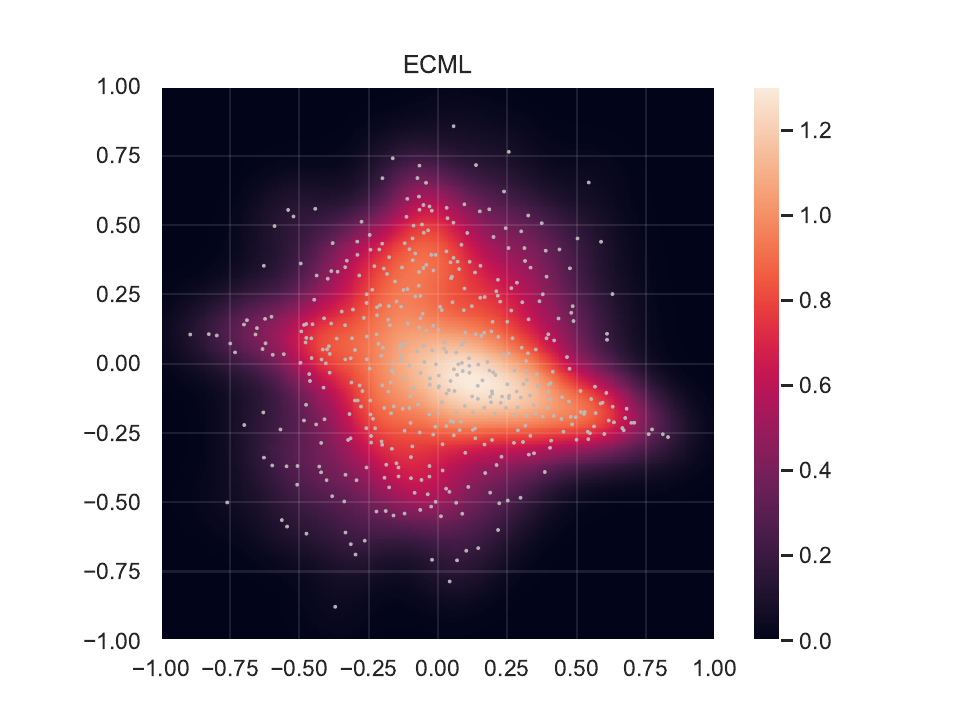}
\includegraphics[width=0.5\columnwidth, trim={1cm 0.5cm 1.5cm 0.5cm}, clip]{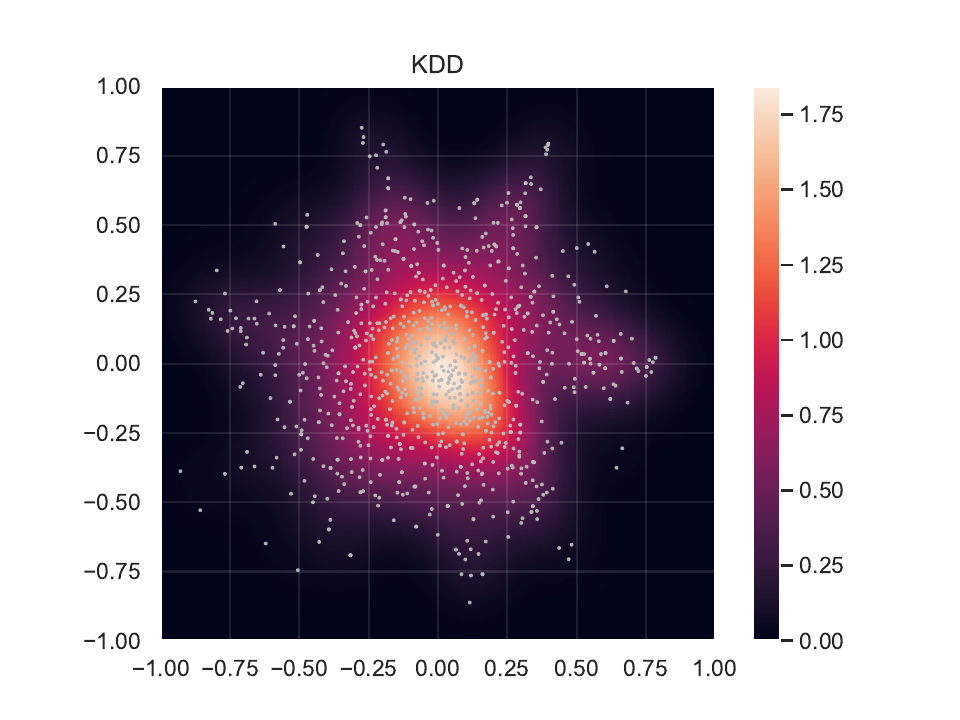}
\includegraphics[width=0.5\columnwidth, trim={1cm 0.5cm 1.5cm 0.5cm}, clip]{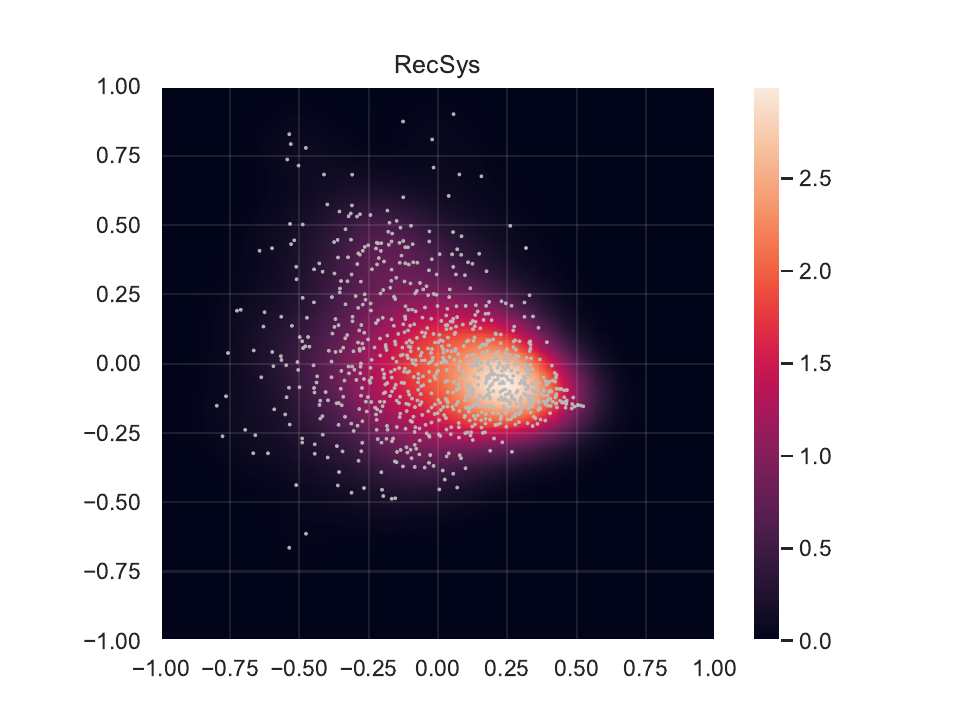}
\includegraphics[width=0.5\columnwidth, trim={1cm 0.5cm 1.5cm 0.5cm}, clip]{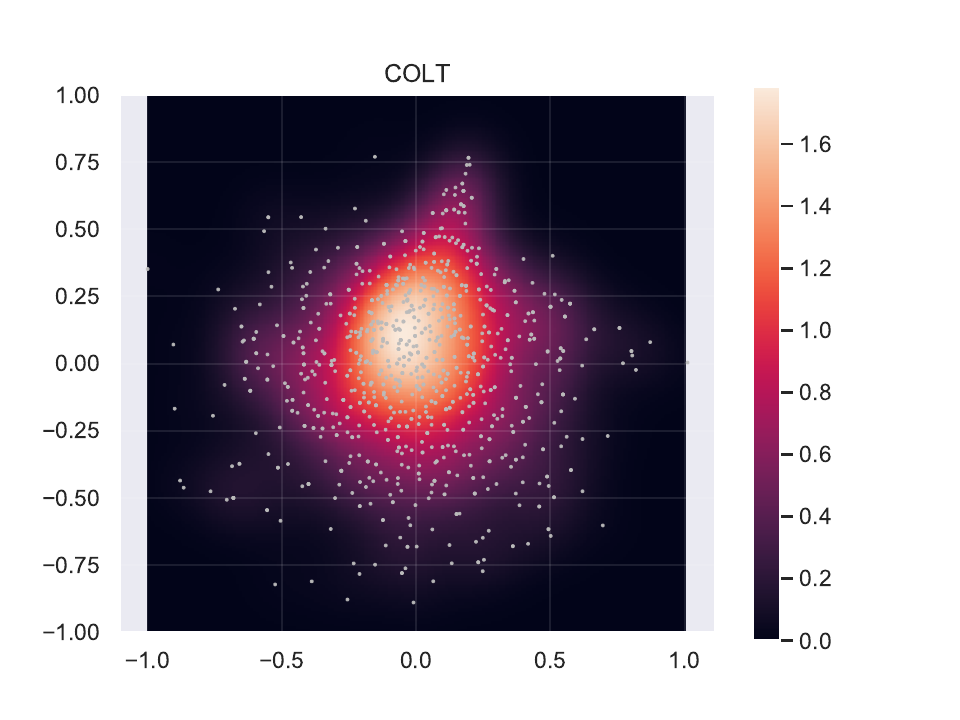}
\includegraphics[width=0.5\columnwidth, trim={1cm 0.5cm 1.5cm 0.5cm}, clip]{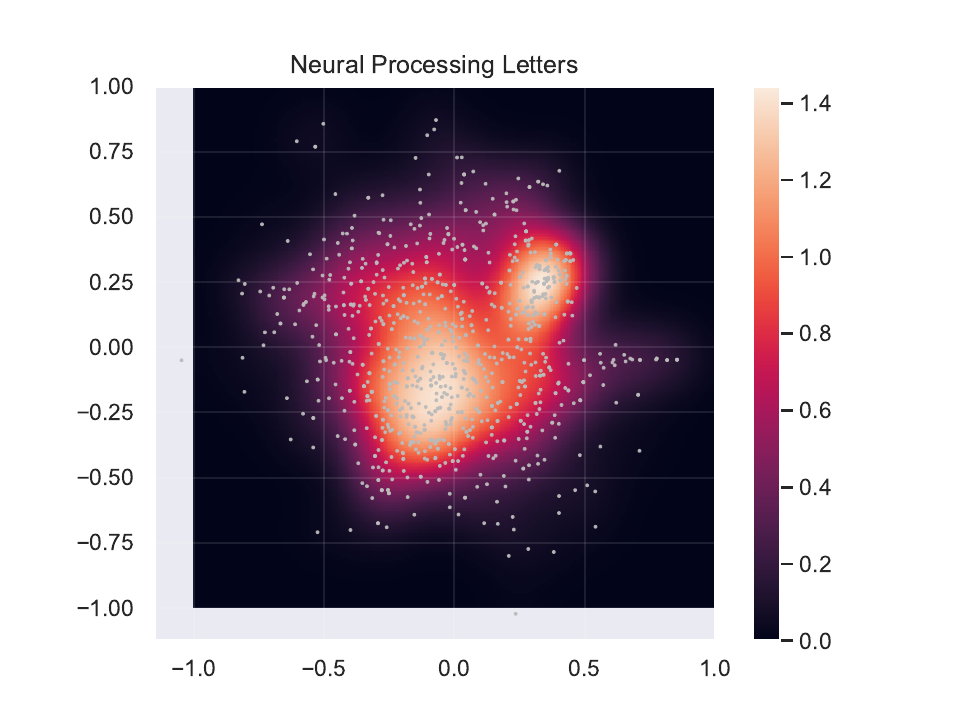}
\includegraphics[width=0.5\columnwidth, trim={1cm 0.5cm 1.5cm 0.5cm}, clip]{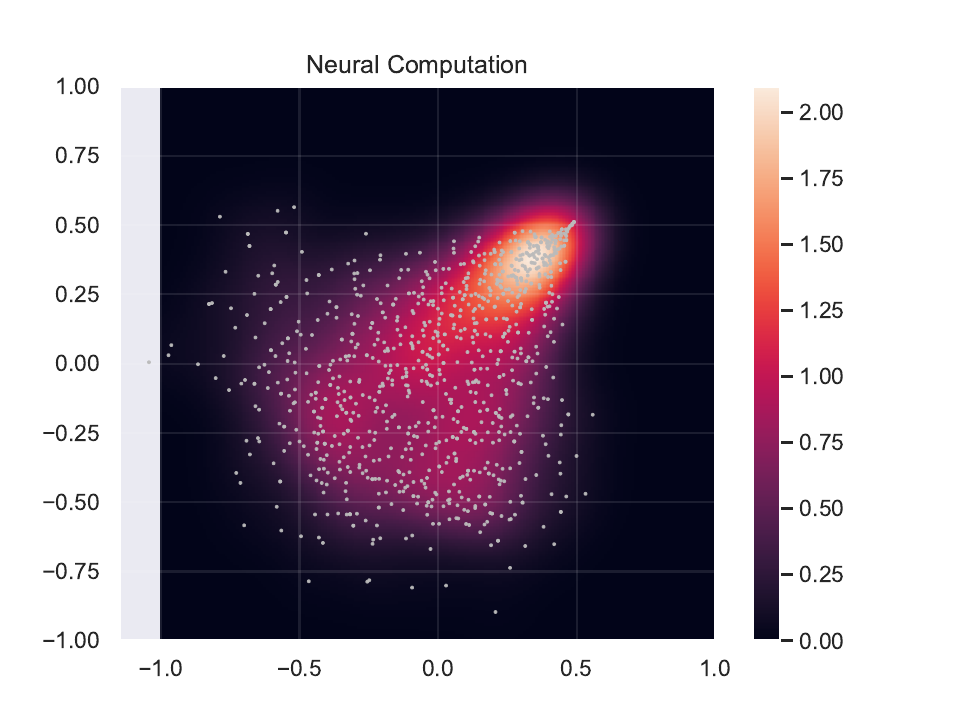}
\includegraphics[width=0.5\columnwidth, trim={1cm 0.5cm 1.5cm 0.5cm}, clip]{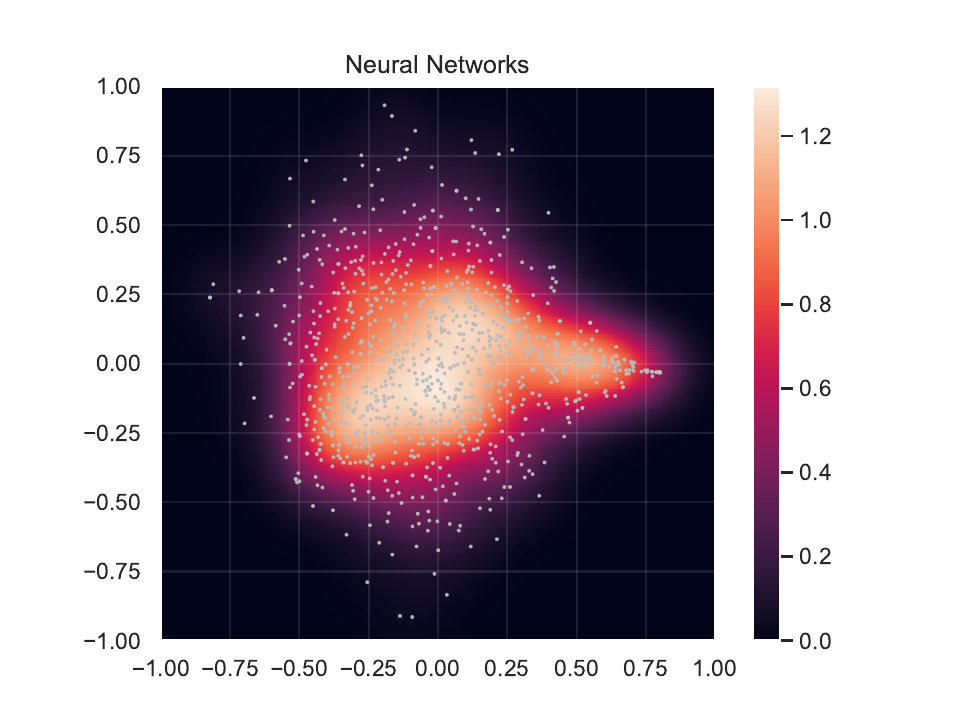}
\includegraphics[width=0.5\columnwidth, trim={1cm 0.5cm 1.5cm 0.5cm}, clip]{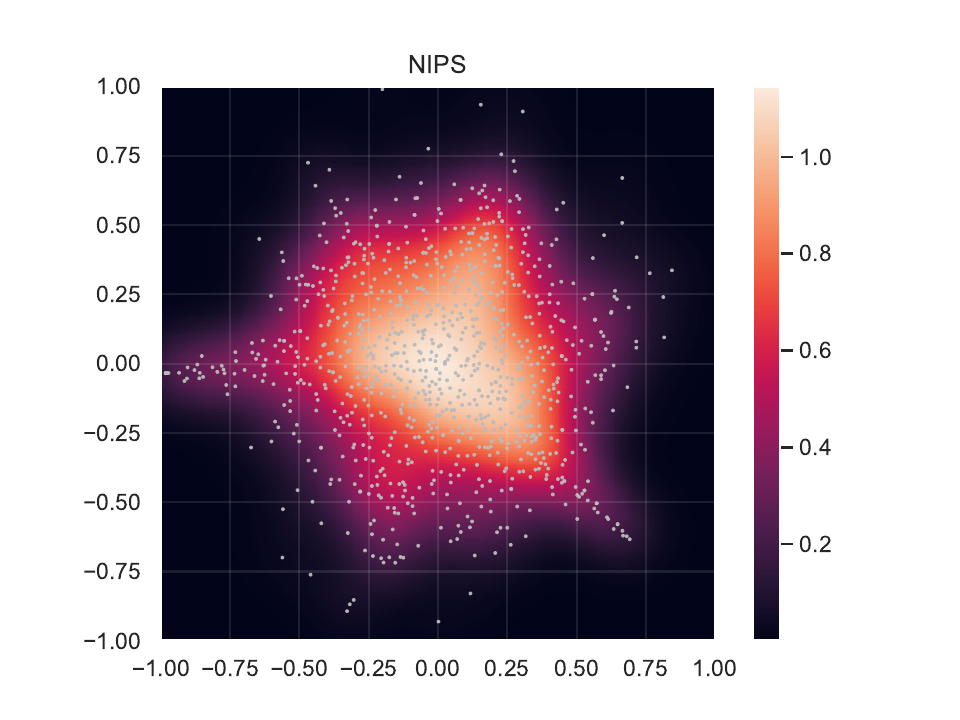}
\caption{Topic densities of selected venues. Dots represent papers after dimensionality reduction of topic vectors through MDS. Proximity in this space hence indicates topical similarity. Colors indicate the probability density of papers estimated through KDE.}\label{fig:topicdensity}
\end{figure}
In this part, we analyze the topical density of some venues. For this, we use the topic space representations of the documents in our data set. For one venue, we project the topic space representations of papers into two-dimensional space using MDS. If we have more than 1000 papers for a venue, we only utilize a random sample of 1000 papers. We do this to speed up our calculations, since MDS has a complexity of $O(n^2)$ for calculating the distances between the sample pairs. After MDS, we do a kernel density estimation (KDE) to estimate a probability density of papers in the two-dimensional space. The KDE is performed with a gaussian kernel and a grid search for the optimal bandwidth, a hyperparameter of this method. Finally, we plot the density as a heat map together with the locations of the projected papers.

While the dimensions in the projected space are not directly interpretable as topics, we can interpret distance in this space as topical distance. An advantage of representing conferences by their distribution over only their centroid is, that we can see the whole distribution of topics instead of only an aggregated mean value. This distribution captures additional information such as various topical hot spots, i.e., dense areas in the projected space.

Figure \ref{fig:topicdensity} shows the results of this process for some selected conferences. We notice here that venues with a broader focus (ECML, KDD) tend to have several ``blobs'' at the margin of the distribution. We suspect that these blobs are all from different topical focusses. Conferences with a stronger single topical focus (COLT, RecSys) do not exhibit this behavior or only slightly. For RecSys, we have a considerably dense area at the right part, while to the left of this papers drift apart from each other. This indicates one single very strong topical focus. Papers deviating from this focus are distributed evenly across other topics, i.e., without any further strongly visible cluster.

The lower four plots all show venues with research on neural networks. In our heat maps in Figure~\ref{fig:confheatmaps} we showed that their topic space trajectories  exhibit similar patterns. In the density plots, however, we see that nonetheless their distributions are different. Topic densities hence reveal additional information to the topic space representations of venues. This is because the venue representations only give an average of the paper vectors in topic space instead of the full distribution. For Neural Processing Letters, two distinct topical focusses are visible, one being broader (i.e., with more variance) than the other one. Neural Computation has a clear focus, similar to RecSys. This is backed by the fact, that it has the second lowest topic diversity directly after RecSys (cf. Table~\ref{tab:confentropy}). NIPS has a broad distribution and papers are focussed on different topics, visible through blobs distributed around the margin. Neural Networks, similar to Neural Processing Letters, seems to have two main focusses. However, there is an area, where these two clusters almost merge. This indicates that there is a smooth thematic transition between both focusses. Hence, we suspect that these blobs are the two topics on neural network.

We saw in this section that topic densities give supplementary information to topic space embeddings. However, in the resulting space, we do not know which area of a density belongs to which topic. At this point, we leave the development of a method for a more specific topical interpretation of areas open for further research. One possibility here would be to create a ``pseudo-paper'' vector for each topic, with full weight on the topic. One could then project these pseudo-papers into the two-dimensional space together with the real papers and mark their positions. It might also be beneficial to apply MDS to the papers of several venues together before estimating the density of each. Finally, we envision that topic space trajectories are extended to such densities, i.e., the chronological development of densities is investigated.

\section{Related Work}
\label{sec:related}

We discussed topic models in more detail in section~\ref{sec:docrepr}. Hence, we
only give a brief outline here. Topic models generally take a document-word
matrix as input, in which documents are represented in a vector space model. The
classic topic model is Latent Semantic Analysis (LSA) \citep{deerwester90lsa},
which performs a singular value decomposition of this matrix. It has been used
for various text mining tasks, e.g., recently in recommendations of scientific
articles \citep{lsarecommendations}. Similarly to LSA, non-negative matrix
factorization (NMF) \citep{Lee99MF} decomposes a matrix into two factors,
however with the constraint that all parts which contribute to the
reconstruction of the input matrix must be positive. The decomposition is
therefore better understandable for humans than through LSA, where topics and
words can also contribute negatively. Latent Dirichlet allocation (LDA)
\citep{lda} is a probabilistic approach on topic modelling that has gained much
interest and is probably the de-facto standard in topic modelling. Many
extensions exist, for example dynamic topic models, where topics change over
time \citep{dlda, blei_lafferty07, topicsovertime} and which incorporate word
embeddings \citep{bleiTopic}. For LDA, some research has reported that it cannot
handle short documents well \citep{Hong2010}.
In our work, however, we look at (relatively short) paper
abstracts. Additionally, in \cite{topicstability} it was shown that LDA produces
considerably less stable results than NMF in repeated runs of the algorithm,
hence leading to impoverished reproducibility. Therefore we refrain from LDA and
related methods in our research. Incorporating dynamic topics as well as word
embeddings complicates interpretability (since topics change over time) and
leads to even worse reproducibility than LDA alone. The reason for the latter is
that these models introduce additional elements of randomness. Hence, we also refrain from methods from these realms. Instead, we utilize
the well-investigated NMF, which gives interpretable and reproducible results.

\added{Review 1.1, 1.2, 2.1, 2.3}{
The work \cite{nmfsurvey} gives an overview over research on NMF. As the authors point out, research on NMF algorithms is focussed on first, applications to different kinds of data, second, enhancing or improving aspects of the algorithm (such as convergence and space and runtime complexity) and third, analyzing algorithmic properties (e.g., concerning the training convergence). In \cite{nmfkld}, it was found that normalizing input data and minimizing Kullback-Leibler (KL) divergence instead of the Frobenius norm may lead to faster convergence and more better approximations. Following this line, \cite{hien2020algorithms} recently proposed new algorithms for KL based NMF with guaranteed non-increasingness of the objective function. In \cite{nmfplsa} it has been shown that with KL divergence, NMF minimize the same objective function as the topic model probabilistic latent semantic indexing (PLSA), which is regarded as a predecessor of LDA. Different optimization methods have been used for NMF, namely based on multiplicative update rules, gradient descent or alternating least squares \citep{nmfsurvey}. The minimization of NMF is often done starting from randomly initialized matrices, which can greatly affect the results \citep{nmfsurvey}. Other initialization methods of the matrices have thus been proposed, e.g., based on a nonnegative singular value decomposition \citep{nndsvd}. 
The work \cite{zhao2016online} proposes an online algorithm for NMF, which is able to learn and update topics without having to store the documents. It is also shown to have good convergence properties as well as being able to handle outliers. The gensim implentation we use in this work is based on the same paper.
By the same authors in \cite{conf/aistats/ZhaoTX17}, NMF is extended to optimize a broader class of divergences as their target function. In this work, the performance, stability and convergence speed in the application as a topic model is demonstrated in a document clustering task, in which documents are assigned to the topic with the strongest weight.
Many further attempts have been made to improve certain properties of NMF or to deal with certain implementation problems. These are often very specific, e.g. to the used optimization algorithm or target function.} 

\added{Review 2.2}{Different attempts have been made to deal with dynamics over time in topic analysis. Generally, there are two basic approaches: 1. Dynamic topic models, which explicitly incorporate time. 2. Using static topic models or word frequencies and to comparing results over time in a post-processing step. 
Examples of the first approach include the aforementioned dynamic topic model (D-LDA), an extension of LDA in which topic distributions exist for time steps and are conditioned on their previous step \citep{dlda}. In \emph{Topics over Time} (TOT) by \cite{topicsovertime}, each topic is modelled as a distribution, which generates words as well as continuous timestamps. This is achieved by extending LDA with an observable timestamp. More recently, the dynamic embedded topic model (D-ETM) has been proposed \citep{bleiTopic}. In this, D-LDA is extended with the recently popular word embeddings. 
Despite the interestingness of these approaches, they also come with some drawbacks in their practical application: The additional parameters lead to more complicated training with often unstable results, which has already been shown for traditional LDA in comparison to NMF  \citep{topicstability}. Furthermore, interpretability is often reduced. This is especially the case when topics (as word distributions) may change over time and need to be reinterpreted.
A different approach is thus to employ static topic models or word frequencies and to compare results over time in a follow-up step. This typically involves sorting results for time steps or slices, sometimes followed by some form of aggregation (as done in our approach) and, finally, an analysis of the results over time. Often, only two slices are employed to provide a before/after comparison in topics and maintaining overview. This, however, may often be to coarse-grained for a deeper analysis.
In Scientometrics, topics are often analyzed by frequent terms in articles or through keywords instead of using explicit topic models. Keywords can be added by authors to their papers to facilitate automatic indexing, e.g., for search engines. 
A field sometimes using this approach and related to our work is the one of \emph{burst detection}, i.e., the detection of sudden increases or decreases of interest in a topic \citep{kleinberg02bursty}. As a more recent application \cite{journals/scientometrics/TattershallNS20} applies burst detection methods from stock market trend analysis to detect bursts in research topics based on normalized term frequencies.
A disadvantage over our approach here is the term-based analysis, which requires additional manual work and does not give an overall summary.
Another popular tool incorporating burst detection as well as bibliometric techniques is CiteSpace \citep{chen_citespace_2006}.
Emerging trend detection here is performed through terms which rapidly grow in frequency over time. Visualizations of scientific research is enriched based on citation and co-citation data. The topical analyses are, however, very coarse-grained. VOSviewer is a more recent tool for the visualization of scientific research \citep{vosviewer}. VOSViewer is based on \emph{Visualization of Similarities} (VOS), a method inspired by and improving on multidimensional scaling when applied to e.g., bibliometric data \citep{vos}. VOSViewer allows the discovery of topics in terms of clustered co-occurring terms or keywords. Comparisons over time can be achieved by comparing visualizations derived from different time slices. VOSViewer is still popular and has been used, e.g., for the analysis of machine learning research in \cite{journals/ijufks/EckW07} and more recently in \cite{salientml}.
Scholia \citep{scholia} is a more recent tool based on Wikidata. It allows to visualize topics of scientific publications based on keywords.
All these approaches do not allow for a fine-grained, topical analysis over time. In contrast to our approach, they do not give an overall summary and comparison over all topics on the venue level and are sometimes difficult to interpret.}

There exists much research where textual analysis of scientific articles has been performed, often involving topic models. In \cite{Mimno2012}, 24 journals from philology and archaeology were analyzed regarding paper locations in a vector space over time and regarding topical variation.
In \cite{griffithss04} article abstracts from the \emph{PNAS} journal were
analyzed for hot topics by performing a linear trend analysis on the weights in
a topic model. In \cite{topicsovertime}, articles from NIPS conference were
analyzed with a dynamic, continuous extension of LDA. The authors showed that
through this method, the increasing and decreasing popularity of recurrent
neural networks during the 1990s could be recovered\added{Review 2.6}{, similar to the findings in our work}. In this work full paper
texts were used, in contrast to our work where abstracts are employed. In \cite{tempcorpus}, based on PNAS articles, a temporal summary consisting of landmark documents, authors and topics was generated. Additionally, in this work papers were layouted in a two-dimensional space based on paper similarities, bibliographic coupling and force-directed layouting. The method for this, called \emph{Vxinsight}, was presented in \cite{vxinsight}. In \cite{blei_lafferty07} and \cite{bleiTopic} the authors demonstrated the applicability of extensions of LDA on articles from the \emph{Science} magazine and the \emph{ACL} conference (Annual Conference of the Association for Computational Linguistics). In \cite{skupin2004:theworld} paper abstracts from geographic research were visualized based on a vector space model, a self-organizing map and hierarchical clustering. 
What all of the above attempts have in common, is that they do not present a thorough, full analysis and comparison of different venues. Additionally, these previous works operate on the paper level, i.e., do not find representations of conferences or journals in vector space. The only exception from this is \cite{Mimno2012}, which, however, solely operates in a dimensionality-reduced space obtained from a bag-of-words representation instead of an interpretable topic space. In contrast to this, we envision methods to analyze topic trajectories of conferences and journals in an interpretable and reproducible manner.

 \added{Review 2.6, 1.5}{Comparing our results to existing scientometric analyses of machine literature, we often found confirmation and sometimes differences. In \cite{journals/ijufks/EckW07}, articles from the computational intelligence field, which strongly overlaps with machine learning, were analyzed using VOS. Some of the clusters they found are similar to our findings, namely \emph{control problems, classification, regression} and \emph{optimization}. A surprising result of this work different from ours, was that clustering and classification were determined as a single topical cluster. \cite{journals/scientometrics/TattershallNS20} analyze computer science research through burst detection and found some results similar to ours, e.g., the rise of popularity of Social Networks and related terms between 2004 and 2014, the lowered interest in \emph{web 2.0} in recent years. In \cite{salientml}, an analysis of machine learning literature has been performed based on measures from social network analysis, VOSviewer and keywords from the web of science. A frequency ranking of topics for 2018 was performed, which was very different from the popularity we found based on our topics. As an example, support vector machines were found to be very popular, which is a contrast to our results as well as our personal background knowledge. An attempt of clustering keywords was made, which, in contrast to our topics, produced clusters of often very different keywords. Finally we want to emphasize that, to the best of our knowledge, we are the first to provide such a deep topical analysis of a large corpus of machine learning literature over time in a long period and on the venue level.}

\section{Conclusion}
\label{sec:conclusion}
In this work we introduced \emph{topic space trajectories}, a novel
approach to analyze conferences and journals. Loosely following the
notion of \emph{interpretable AI}, our focus here lay on interpretable results and reproducible methods.  \added{Review 1.8}{Overall, additionally to introducing our method, our aim was to present a deep analysis of the results achievable through it.} We therefore demonstrated our approach on a set of machine learning conferences and journals and came to various interesting insights. Additionally to being reasonable and fascinating by themselves, the presented results support the applicability of our approach. Nonetheless, we also found some limitations. Most of these limitations stem from the nature of high-dimensional data. The reduction of high-dimensional documents in a \emph{tf-idf} representation to a low-dimensional topic space representation mitigates such problems. However, the topic space still has a dimension that cannot be visualized in a coordinate system. As a consequence, we further reduced our data to a two-dimensional space through multidimensional scaling. In this representation we can analyze topical (dis)similarities between objects (i.e., documents or venues) in what we called \emph{topical maps} and \emph{topic densities}. However, we obtain dimensions that are not interpretable as topics anymore and lose information about \emph{which} topics are (dis)similar. We therefore found salvation in two solutions: First, plotting trajectories for only the two most interesting topics, and second, visualizations of trajectories as heat maps. Based on topic space trajectories, we also analyzed topical diversities of venues and their development over time. Last but not least, we made our discoveries plausible throughout our analysis through a comparison with historical events in research.

Many of our approaches open room for further research. An interesting direction would be automated name suggestions for the topics found through NMF. We imagine that this could be achieved through frequent bigrams in the most relevant documents for a topic. Topical maps and topic densities could be extended by their development in time, i.e., by calculating them for trajectories. Both would benefit from methods for improved interpretability.

As already indicated in our introduction, we envision further applications of our method. First, our venue embeddings could be used in a recommendation scenario, e.g., similar as in \cite{lsarecommendations}. Second, it would be interesting to explore whether topic space trajectories can be extrapolated into the future. This could then, as an example, lead to even better recommendations. Possible applications lie in researchers looking for a suitable conference or journal for an unpublished paper or for recommending reviewers for a submitted paper. Last, we are eager to see our method applied to other kinds of data. As one very similar example, instead of venues we could analyze the trajectories of authors in topic space. Our method can, however, also be applied to totally different text domains.

\begin{acknowledgements}
  This work was funded by the German Federal Ministry of Education and
  Research (BMBF) in its program ``Quantitative
  Wissenschaftsforschung'' as part of the REGIO project under grant
  01PU17012 as well as the German Research Foundation (DFG) priority
  programme (SPP) 1894 project topikos.
\end{acknowledgements}

\bibliographystyle{spbasic}
\bibliography{paper}
\end{document}